\pdfoutput=1
\PassOptionsToPackage{dvipsnames,table,xcdraw}{xcolor}

\documentclass[11pt]{article}
\usepackage{booktabs}  
\usepackage{multirow}  
\usepackage{multicol}
\usepackage{amsmath}
\usepackage{amssymb}
\usepackage{bm} 
\usepackage{graphicx}
\usepackage{placeins}
\usepackage{float}
\usepackage{tcolorbox}
\usepackage{enumitem}
\usepackage{colortbl}
\usepackage{caption}
\usepackage{xcolor}
\definecolor{LimeGreen}{rgb}{0.196,0.804,0.196} 
\usepackage[final]{acl}

\usepackage{times}
\usepackage{latexsym}
\usepackage[T1]{fontenc}

\usepackage[utf8]{inputenc}

\usepackage{microtype}

\usepackage{inconsolata}

\title{\textsc{CoCoA}: \underline{Co}nfidence- and \underline{Co}ntext-Aware \underline{A}daptive Decoding for Resolving Knowledge Conflicts in Large Language Models}

\author{Anant Khandelwal, Manish Gupta, Puneet Agrawal \\
  Microsoft, India\\
  \texttt{\{anantk,gmanish,punagr\}@microsoft.com} \\
}

\definecolor{hallucination}{RGB}{200,0,0}  
\definecolor{synthesis}{RGB}{229,151,0}    
\definecolor{headergray}{gray}{0.9}

\begin{document}
\maketitle
\begin{abstract}
Faithful generation in large language models (LLMs) is challenged by knowledge conflicts between parametric memory and external context. Existing contrastive decoding methods tuned specifically to handle conflict often lack adaptability and can degrade performance in low conflict settings. We introduce \textsc{CoCoA} (\underline{Co}nfidence- and \underline{Co}ntext-Aware \underline{A}daptive Decoding), a novel token-level algorithm for principled conflict resolution and enhanced faithfulness. 
\textsc{CoCoA} resolves conflict by utilizing confidence-aware measures (entropy gap and contextual peakedness) and the generalized divergence between the parametric and contextual distributions.
Crucially, \textsc{CoCoA} maintains strong performance even in low conflict settings. Extensive experiments across multiple LLMs on diverse Question Answering (QA), Summarization, and Long-Form Question Answering (LFQA) benchmarks demonstrate \textsc{CoCoA}'s state-of-the-art performance over strong baselines like \textsc{AdaCAD}. It yields significant gains in QA accuracy, up to 9.2 points on average compared to the strong baseline \textsc{AdaCAD}, and improves factuality in summarization and LFQA by up to 2.5 points on average across key benchmarks. Additionally, it demonstrates superior sensitivity to conflict variations. \textsc{CoCoA} enables more informed, context-aware, and ultimately more faithful token generation.
\end{abstract}

\section{Introduction}
Large language models (LLMs) have achieved strong performance across Natural Language Processing (NLP) tasks such as question answering, summarization, and fact verification by leveraging vast parametric knowledge acquired during pretraining ~\citep{petroni-etal-2019-language, roberts-etal-2020-much, NEURIPS2020_1457c0d6}. However, this knowledge is static and limited by the training data, making it prone to becoming outdated or incomplete ~\citep{10.5555/3540261.3542508, sun-etal-2024-head, hernandez2024inspectingeditingknowledgerepresentations}. To address this, \textit{context-aware generation} augments LLMs with auxiliary inputs (such as retrieved documents or tool outputs) at inference time, enabling incorporation of up-to-date, task-specific knowledge without retraining ~\citep{10.5555/3524938.3525306, nakano2021webgpt, 10.5555/3666122.3669119}. Yet, this introduces the risk of \textit{knowledge conflict} between the external context $\bm{c}$ and the model’s internal knowledge ~\citep{chen-etal-2022-rich, xie2024knowledgeconflict, li-etal-2023-large}, especially when the two sources contradict each other.

\begin{figure*}[h]
    \centering
    \includegraphics[width=\textwidth]{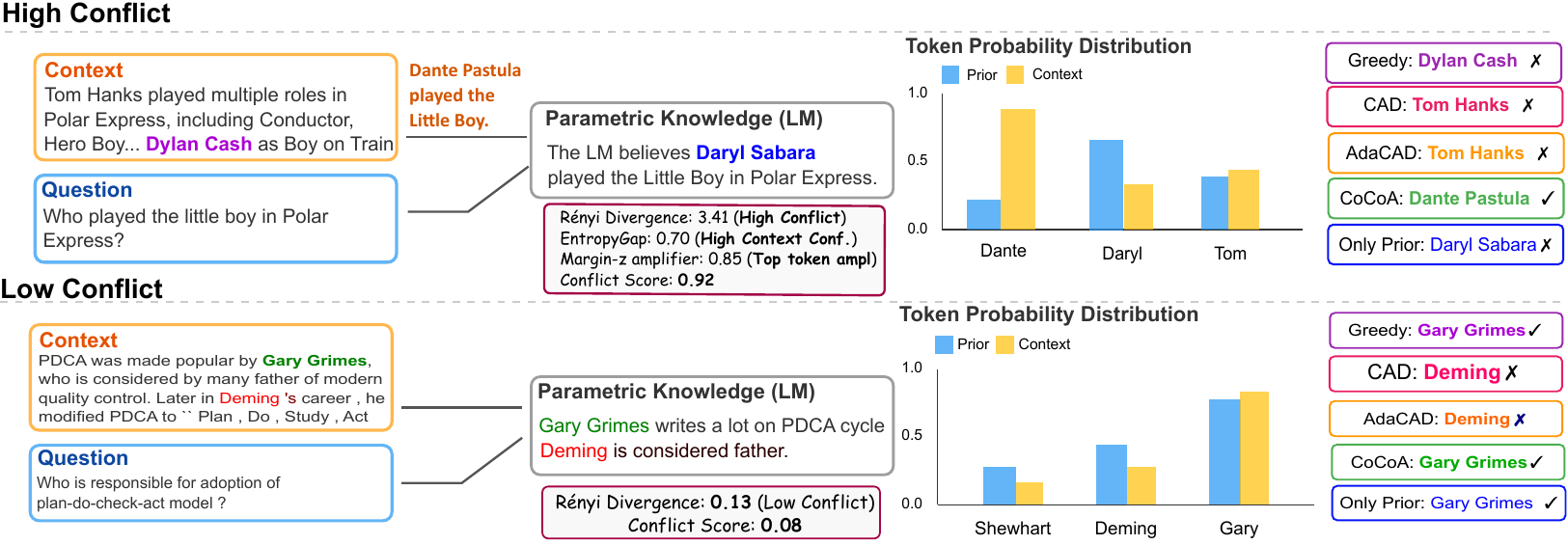}
    \caption{\textit{Comparison of decoding methods on high- and low-conflict questions. \textsc{CoCoA} accurately resolves high-conflict answers by leveraging confidence-aware conflict resolution and preserves correct predictions in low-conflict cases, outperforming greedy, CAD, and AdaCAD.}}
    \label{fig:CoCoA_arch}
\end{figure*}

Standard decoding methods over the context-conditioned distribution $p_{\theta}^{\text{ctx}}$ often fail to resolve such conflicts, defaulting to parametric priors even when contradicted by contextual evidence ~\citep{longpre2021entity, zhou-etal-2023-context}. This \textit{model stubbornness} undermines performance in knowledge-sensitive settings (Fig.~\ref{fig:CoCoA_arch}(top)). \textit{Context-aware decoding} methods attempt to mitigate this by contrasting $p_{\theta}^{\text{ctx}}$ with the unconditional distribution $p_{\theta}$, promoting tokens grounded in the context and penalizing those favored only by internal memory. For instance, \textbf{Context-aware Decoding (CAD)} ~\citep{shi-etal-2024-trusting} applies a fixed contrastive parameter $\alpha$ to reweight token probabilities. While CAD improves outcomes under strong conflict, its static nature can lead to \textit{over-correction} when context and model agree, and \textit{under-correction} in subtle conflicts, harming output quality in low-conflict cases ~\citep{wang2023resolving}.

\textsc{AdaCAD} ~\citep{wang-etal-2025-adacad} extends CAD by dynamically adjusting contrastive weights at each decoding step using the Jensen-Shannon Divergence (JSD) between $p_{\theta}^{\text{ctx}}$ and $p_{\theta}$, removing the need for a fixed hyperparameter. However, it inherits two key limitations. First, JSD uniformly penalizes all distributional shifts, failing to distinguish meaningful context signals from noise. 
Second, it saturates on peaked or heavy-tailed distributions typical in autoregressive decoding, limiting sensitivity to subtle conflicts, especially early in generation. AdaCAD mitigates this with heuristic ``warm-up'' inflation, introducing new hyperparameters and reducing adaptability.

To address these issues, we propose \textsc{CoCoA} (\underline{Co}nfidence- and \underline{Co}ntext-Aware \underline{A}daptive Decoding), a principled framework (in Fig.~\ref{fig:CoCoA_arch}) for resolving knowledge conflicts in context-aware generation at test time. 
Our key contributions are as follows:
(a) \textbf{Tail-Sensitive Conflict Detection}: We replace JSD with Rényi divergence, enabling tunable sensitivity across entropy regimes. This enhances the model’s ability to detect tail-heavy shifts between prior and contextual distributions, crucial for capturing subtle but meaningful conflicts.
(b) \textbf{Contextual Confidence Estimation}: We introduce a novel confidence signal that combines the entropy gap between prior and context distributions with the contextual peakedness, capturing both divergence and certainty in a unified measure.
(c) \textbf{Adaptive Gating Mechanism}: We propose a dynamic gating strategy that determines how much to trust the context at each step, leveraging a stable blend of conflict signals and contextual confidence. 
Experiments on diverse context-rich tasks (including QA, long-form generation, and fact verification) show that \textsc{CoCoA} achieves state-of-the-art accuracy compared to other test-time baselines while maintaining fluency and resolving conflicts effectively. Code: \url{https://github.com/infusion-zero-edit/CoCoA/}.

\section{Related Work}
\paragraph{Context-aware Decoding and Knowledge Integration.} 
Recent studies tackle misalignment between retrieved context and parametric knowledge, especially in entity-centric QA ~\citep{longpre2021entity} and retrieval-augmented generation ~\citep{tan-etal-2024-blinded}. While external knowledge can reduce hallucinations ~\citep{shuster-etal-2021-retrieval-augmentation}, balancing it with model priors remains challenging. Some approaches rely on prompt engineering ~\citep{zhou-etal-2023-context} or train auxiliary discriminators ~\citep{zhang-etal-2023-merging}, whereas our method is training-free and prompt-agnostic.
\paragraph{Contrastive Decoding Methods.} 
Contrastive decoding enhances contextual grounding ~\citep{shi-etal-2024-trusting}, diversity ~\citep{li-etal-2016-diversity}, and controllability ~\citep{liu-etal-2021-dexperts}. CAD ~\citep{shi-etal-2024-trusting} contrasts contextualized vs. unconditional outputs, while ConfCD ~\citep{zhao-etal-2024-enhancing} adjusts generation confidence with noisy inputs. AdaCAD ~\citep{wang-etal-2025-adacad} introduces a continuous conflict spectrum using JSD, avoiding discrete binning.
\paragraph{Faithfulness in Long-Form QA (LFQA).}
LFQA systems often hallucinate beyond retrieved evidence ~\citep{fan-etal-2019-eli5, han-etal-2024-rag}. Techniques like RECOMP ~\citep{xu2024recomp} and SelfRAG ~\citep{asai2024selfrag} aim to improve relevance via filtering or reflection during training. Contrastive methods ~\citep{li-etal-2023-contrastive, shi-etal-2024-trusting, wang-etal-2025-adacad} enhance context salience, but often rely on post-hoc steps. Our method improves contextual alignment during generation without extra decoding passes.
\paragraph{Knowledge Conflict in LLMs.} 
Conflicts between contextual and parametric knowledge, i.e., context-memory conflicts ~\citep{chen-etal-2022-rich, xie2024knowledgeconflict} cause models to ignore retrieved facts. ~\citet{xu-etal-2024-knowledge-conflicts} categorize conflict types; we target dynamic resolution of context-memory conflict. Unlike revision-based approaches ~\citep{10.1145/3703155, choi-etal-2023-kcts}, our method integrates conflict signals directly into decoding for test-time correction.
\section{Methodology}
\subsection{Task Setup and Notation}
We consider autoregressive generation tasks (e.g., QA, instruction following), where a model generates a sequence $\bm{y} = (y_1, \dots, y_T)$ given a query $\bm{x}$ and optional context $\bm{c}$. The model defines two token distributions: $p_\theta(y_t \mid \bm{x}, \bm{y}_{<t})$ (prior) and $p_\theta^{\mathrm{ctx}}(y_t \mid \bm{c}, \bm{x}, \bm{y}_{<t})$ (contextualized). Standard decoding treats context as static input, ignoring conflicts between $p_\theta$ and $p_\theta^{\mathrm{ctx}}$. Our goal is to resolve such conflicts during generation by amplifying trustworthy context while avoiding overreliance on either source.
\subsection{Background: \textsc{AdaCAD}}
\textsc{AdaCAD} ~\citep{wang-etal-2025-adacad} adapts contrastive decoding by replacing the fixed weight $\alpha$ in CAD ~\citep{shi-etal-2024-trusting} with a dynamic, stepwise signal. At each token $t$, it computes the JSD between the context-aware and prior distributions:
\begin{equation}
    \small\alpha^{\text{JSD}}_t = \text{JSD}\left(p_\theta^{\text{ctx}} \,\|\, p_\theta\right)
\end{equation}
and samples from a blended distribution:
\begin{equation}
    \small\tilde{p}_\theta \propto p_\theta^{\text{ctx}} \cdot \left( \frac{p_\theta^{\text{ctx}}}{p_\theta} \right)^{\alpha_t^{\text{JSD}}}
\end{equation}
This lets \textsc{AdaCAD} increase reliance on context when conflicts are high and defer to the model prior when they agree, offering token-level adaptability absent in CAD.
\subsection{\textsc{CoCoA}: Confidence- and Context-aware Adaptive Decoding)}
We propose \textsc{CoCoA} (Confidence- and Context-aware Adaptive Decoding), a token-level decoding algorithm that dynamically resolves conflicts between a model's parametric knowledge and external context. 
\textsc{CoCoA} resolves conflict by utilizing confidence-aware measures (entropy gap and contextual peakedness) and the generalized (Rényi) divergence between the prior and contextual distributions. Harnessing Rényi divergence and entropy gap in \textsc{CoCoA} enables it to better capture subtle shifts between the two distributions, especially in low-confidence regimes (Fig.\ref{fig:CoCoA_arch}). At each decoding step \( t \), \textsc{CoCoA} considers two distributions over the next token: the base model’s prior \( p_\theta(y_t) \) and the context-aware version \( p_\theta^{\text{ctx}}(y_t) \). Rather than favoring one arbitrarily, \textsc{CoCoA} blends them via a conflict-aware weight \( \lambda_t \in [0,1] \), producing:
\begin{equation}
    \small q(y_t) \;\propto\; p_{\theta}(y_t)^{1-\lambda_t} \cdot p_{\theta}^{\text{ctx}}(y_t)^{\lambda_t},
\end{equation}
which corresponds to logit interpolation:
\begin{equation}
    \small\hspace{-8pt}\log q(y_t) = (1-\lambda_t)\log p_{\theta}(y_t) + \lambda_t \log p_{\theta}^{\text{ctx}}(y_t),
\end{equation}
followed by renormalization. The interpolation weight \( \lambda_t \) is dynamically computed based on the measured conflict between the two distributions, with Rényi divergence providing tunable sensitivity across entropy regimes.
\subsubsection{Conflict Detection in \textsc{CoCoA}}
To detect conflicts between the prior distribution \( p_{\theta}(y_t) \) and the context-aware distribution \( p_{\theta}^{\text{ctx}}(y_t) \), \textsc{CoCoA} leverages two complementary signals: Rényi divergence and entropy gap.

We first compute the \textit{Rényi divergence} of order \( \alpha \), a generalization of KL divergence that is sensitive to discrepancies in the tail of distributions. For \( \alpha \neq 1 \), the divergence is defined as:
\begin{equation}
\hspace{-5pt}\small
    D^{\alpha}_{t}(p_\theta \Vert p_\theta^{\text{ctx}}) = \frac{1}{\alpha-1} \log \sum_{i=1}^{|V|} p_\theta(y_t^{(i)})^\alpha \, p_\theta^{\text{ctx}}(y_t^{(i)})^{1-\alpha}
\end{equation}
\noindent where $V$ is vocabulary. Choosing \(\alpha<1\) emphasizes low-probability events, allowing the model to surface sharp contextual shifts even when overall token distributions seem similar, unlike symmetric measures such as JSD. To further capture context-induced changes in uncertainty, we compute the \textit{entropy gap}:
\begin{equation}
    \small\Delta H_t = H(p_{\theta}(y_t)) - H(p_{\theta}^{\text{ctx}}(y_t)),
\end{equation}
where entropy is given by:
\begin{equation}
    \small H(P) = -\sum_y P(y) \log P(y).
\end{equation}
A large positive \( \Delta H_t \) indicates that context has increased certainty by concentrating probability mass, while a small or negative gap suggests minimal contextual influence. This helps differentiate between noisy and confident context, e.g., even when divergence is moderate, a high entropy gap signals that the context provides a strong directional cue. Together, Rényi divergence and the entropy gap allow \textsc{CoCoA} to detect not only distributional disagreement but also changes in certainty, enabling finer-grained adaptation to conflict during decoding.
\subsubsection{Contextual Peakedness in \textsc{CoCoA}}
To refine the model's sensitivity to confident contextual cues, we introduce a 
contextual peakedness measure based on the sharpness of the context distribution. Let \( m_t \) denote the margin between the top two token probabilities under the context-aware distribution:
\begin{equation}
    \small m_t = p_{\theta}^{\text{ctx}}(y_t^{(1)}) - p_{\theta}^{\text{ctx}}(y_t^{(2)}),
\end{equation}
where \( y_t^{(1)} \) and \( y_t^{(2)} \) are the highest-ranked and second-ranked candidates, respectively. A large margin suggests high certainty in the context's preferred token. 
This factor ensures that confident contextual signals have stronger influence during decoding.
\subsubsection{Adaptive Gating in \textsc{CoCoA}}
To determine how much to trust the context at each step, we compute a conflict score \( s_t \) based on divergence and entropy signals:
\begin{equation}
    \small s_t = \sigma\left( D^{\alpha}_t + \gamma \Delta H_t + \delta \right),
    \label{mixing_entropy_renyi}
\end{equation}
where \( D^{\alpha}_t \) is the Rényi divergence, \( \Delta H_t \) is the entropy gap, \( \sigma \) is the sigmoid function, $\gamma$ is the mixing weight between \( \Delta H_t \) and \( D^{\alpha}_t \), and \( \delta \) is a small constant for numerical stability. This score reflects the severity of knowledge conflict between the model and the context.

We then integrate the contextual peakedness \( m_t \) with the conflict score to form blending weight:
\begin{equation}
    \small \lambda_t = \sigma\left( z\log m_t + \log \frac{1 - s_t}{s_t} \right), \quad \text{with } z > 1,
\end{equation}
which sharpens the gating decision: confident context (large \( m_t \)) and high conflict (large \( s_t \)) push \( \lambda_t \) closer to 1; low conflict or weak context drive it toward 0. Finally, we compute the blended output distribution using power interpolation:
\begin{equation}
    \small q(y_t) \propto p_{\theta}^{\text{ctx}}(y_t)^{\lambda_t} \cdot p_{\theta}(y_t)^{1 - \lambda_t},
\end{equation}
The \textit{logit-space normalization term} (unlike heuristic ``warm-up'' in CAD) stabilizes blending and preserves generation quality. This formulation allows \textsc{CoCoA} to adaptively modulate reliance on context versus prior, ensuring more trustworthy and calibrated generation across varying levels of knowledge conflict.
\section{Experiments and Results}
\subsection{Experimental Setup}
\paragraph{Datasets and Metrics.}  
We evaluate our approach across a diverse set of benchmarks. For foundational question answering (QA), we use Natural Questions (NQ; ~\citep{10.1162/tacl_a_00276}), TriviaQA ~\citep{joshi-etal-2017-triviaqa}, PopQA ~\citep{mallen-etal-2023-trust}, and HotpotQA ~\citep{yang-etal-2018-hotpotqa}. To assess robustness to conflicting information, we include NQ-SWAP ~\citep{longpre2021entity}, a synthetic conflict variant of NQ. Structured reasoning is tested with the tabular QA dataset TabMWP ~\citep{DBLP:conf/iclr/Lu0CWZRCK23}. Performance on these QA benchmarks is measured by exact match accuracy.

For long-form generation, we use CNN-DM ~\citep{see-etal-2017-get}, XSum ~\citep{narayan-etal-2018-dont}, and the topic-focused dialogue summarization dataset TofuEval ~\citep{tang-etal-2024-tofueval}, which emphasizes marginal topics. Summarization quality on CNN-DM and XSum is evaluated via ROUGE-L ~\citep{lin-2004-rouge} and BERT-P ~\citep{Zhang*2020BERTScore:}. Since TofuEval lacks reference summaries, we measure factual consistency with AlignScore ~\citep{zha-etal-2023-alignscore} for both main and marginal topics.

Our evaluation also covers diverse long-form QA (LFQA) datasets spanning various query types and complexities: CLAPNQ ~\citep{rosenthal-etal-2025-clapnq} (real web queries with gold documents and multi-source simulation), ExpertQA ~\citep{malaviya-etal-2024-expertqa} (expert questions with verified answers), HAGRID ~\citep{hagrid} (information-seeking questions with LLM-generated, evaluated answers), ELI5-WebGPT ~\citep{nakano2021webgpt} (``Explain Like I'm Five'' style with human answers), and QuoteSum ~\citep{schuster-etal-2024-semqa} (semi-extractive answers from multiple sources).

Evaluation metrics are dataset-specific: CLAPNQ, ExpertQA, and HAGRID use ROUGE-L; ELI5-WebGPT employs claim recall between generated responses and gold sub-claims ~\citep{chen2023understandingretrievalaugmentationlongform}; QuoteSum is evaluated with SEMQA ~\citep{schuster-etal-2024-semqa}. Faithfulness (how well responses are grounded in the provided context) is measured by MiniCheck ~\citep{tang2024minicheckefficientfactcheckingllms}, which scores consistency of statements against source documents. We report the average consistency score (FaithScore) per dataset. Additional details and examples are provided in Appendix~\ref{sec:datasets_appendix}.
\paragraph{Source of Context.}  
For all experiments, we use gold contexts provided by each dataset to ensure consistent and standardized evaluation. For NQ, NQ-SWAP, TriviaQA, HotpotQA, and PopQA, we adopt the gold contexts curated by the AdaCAD authors ~\citep{wang-etal-2025-adacad}. In TabMWP, the accompanying semi-structured tables serve as contextual input.

For summarization tasks (CNN-DM, XSum, and TofuEval) the source documents are used as context, with the task-specific prompts serving as input queries. We expect context distributions to be much more confident compared to prior distributions here, and hence the gating parameter $\lambda_t$ should increase with decoding steps.
In the long-form QA setting, we use the gold contexts included with each dataset: curated web documents in ELI5-WebGPT ~\citep{nakano2021webgpt}, expert-verified domain documents in ExpertQA ~\citep{malaviya-etal-2024-expertqa}, retrieved passages for attributed explanations in HAGRID ~\citep{hagrid}, selected passages tailored for long-form answers in CLAPNQ ~\citep{rosenthal-etal-2025-clapnq}, and Wikipedia passages supporting semi-extractive answers in QuoteSum ~\citep{schuster-etal-2024-semqa}. A full summary of input queries \(x\) and associated contexts \(c\) is presented in Table~\ref{tab:dataset-prompts}, with prompt details provided in Appendix~\ref{sec:prompts}.
\paragraph{Models.} We evaluate \textsc{CoCoA} using multiple pretrained models, including LLaMA 2 (13B) ~\citep{touvron2023llama2openfoundation}, LLaMA 3 (8B, 70B) ~\citep{llama3modelcard}, and Mistral (7B) ~\cite{jiang2023mistral7b}, assessing performance on both base and instruction-tuned variants to ensure robustness across model types.
\paragraph{Baselines.}  
We compare \textbf{\textsc{CoCoA}} against a range of established and recent test-time decoding methods all of which also take context as part of input. These include: (1) \textbf{Greedy decoding}, which selects tokens directly from the model’s output without contextual adjustments; (2) \textbf{Context-aware decoding (CAD)} ~\citep{shi-etal-2024-trusting}, which applies a fixed contrastive scaling factor $\alpha$ across all examples; (3) \textbf{COIECD} ~\citep{yuan-etal-2024-discerning}, which classifies each step as conflicting or not using a threshold $\lambda$, switching between fixed decoding modes accordingly; and (4) \textbf{ConfCD} ~\citep{zhao-etal-2024-enhancing}, which dynamically sets $\alpha$ at each timestep based on confidence differences between context-aware and context-free predictions.

We also include \textbf{\textsc{AdaCAD}} ~\citep{wang-etal-2025-adacad}, the strongest baseline method, which adaptively sets $\alpha_t$ using JSD between prior and context-aware distributions to model conflict dynamically. For CAD, we follow prior work with $\alpha = 1.0$ for QA and $\alpha = 0.5$ for summarization. For COIECD, we use $\lambda = 0.25$ and the same $\alpha$ values as CAD. ConfCD sets $\alpha = \max_{y'} p_\theta(y'|c, x, y_{<t})$ if this context-conditioned probability exceeds its context-free counterpart; otherwise, $\alpha = 1 - \max_{y'} p_\theta(y'|x, y_{<t})$.

\textbf{\textsc{CoCoA}} differs by computing adjustment factors via divergence-stabilized entropy ratios and contextual confidence estimation. Following extensive ablation, we fix hyperparameters to $\alpha = 0.5$ (Rényi divergence), $z = 5.0$ (contextual peakedness mixing weight), $\gamma = 1.0$ (entropy gap weight), and $\delta = 1\text{e}^{-8}$. All methods are evaluated using zero-shot greedy decoding to ensure fair comparison across QA and summarization tasks. We used the eval scripts obtained from~\citet{wang-etal-2025-adacad}.
\subsection{Main Results}
\begin{table}[ht]
    \centering
    \tabcolsep1pt
    \scriptsize
    \begin{tabular}{lllccccccc}
    \toprule
    & Decoding & NQ   & NQ-SWAP & TriviaQA & PopQA & HotpotQA & TabMWP & Avg \\
    \midrule
    \multirow{5}{*}{\rotatebox{90}{\textbf{Llama2-13B}}} & Greedy   & 44.26 & 54.89  & 85.50 & 76.65 & 38.27   & 38.30  & 56.31 \\
                         & CAD      & 37.91 & 80.35  & 71.40 & 76.83 & 31.92   & 19.30  & 52.95 \\
                         & COIECD   & 44.60 & 59.84  & 87.00 & 81.05 & 42.81   & 38.80  & 59.02 \\
                         & \textsc{AdaCAD}   & 46.73 & 67.84  & 85.40 & 78.79 & 37.83   & 37.50  & 59.02 \\
                         & \textsc{CoCoA}     & \textbf{49.49} & \textbf{80.36} & \textbf{89.00} & \textbf{87.29} & \textbf{43.04} & \textbf{44.00} & \textbf{65.53} \\
    \midrule
    \multirow{5}{*}{\rotatebox{90}{\textbf{Llama3-8B}}}  & Greedy   & 44.63 & 47.81  & 85.70 & 80.51 & 51.42   & 52.20  & 60.38 \\
                         & CAD      & 35.96 & 77.94  & 40.20 & 74.27 & 39.53   & 26.60  & 49.08 \\
                         & COIECD   & 43.36 & 51.16  & 83.10 & 78.49 & 45.63   & 49.70  & 58.57 \\
                         & \textsc{AdaCAD}   & 45.47 & 62.34  & 82.50 & 81.34 & 50.53   & 53.00  & 62.53 \\
                         & \textsc{CoCoA}     & \textbf{49.30} & \textbf{79.15} & \textbf{90.40} & \textbf{93.76} & \textbf{51.01} & \textbf{62.80} & \textbf{71.74} \\
    \midrule
    \multirow{5}{*}{\rotatebox{90}{\textbf{Llama3-70B}}} & Greedy   & 44.13 & 55.74  & 90.20 & 86.10 & 56.11   & 66.70  & 66.50 \\
                         & CAD      & 34.05 & 81.32  & 54.60 & 75.16 & 40.86   & 48.60  & 55.77 \\
                         & COIECD   & 45.09 & 57.26  & 88.60 & 83.60 & 52.03   & 64.40  & 65.16 \\
                         & \textsc{AdaCAD}   & 45.43 & 70.07  & 88.80 & 85.68 & 55.00   & 67.10  & 68.68 \\
                         & \textsc{CoCoA}     & \textbf{51.80} & \textbf{88.32} & \textbf{93.00} & \textbf{95.90} & \textbf{59.29} & \textbf{78.90} & \textbf{77.87} \\
    \midrule
    \multirow{5}{*}{\rotatebox{90}{\textbf{Mistral-7B}}} & Greedy   & 42.56 & 56.86  & 80.40 & 67.56 & 40.89   & 38.90  & 57.65 \\
                         & CAD      & 20.98 & 66.89  & 24.20 & 48.54 & 18.49   & 20.10  & 35.82 \\
                         & COIECD   & 29.00 & 58.09  & 71.60 & 64.59 & 35.83   & 31.60  & 48.45 \\
                         & \textsc{AdaCAD}   & 45.09 & 67.27  & 80.20 & 67.26 & 41.35   & 39.70  & 60.23 \\
                         & \textsc{CoCoA}     & \textbf{48.00} & \textbf{80.90} & \textbf{87.70} & \textbf{76.83} & \textbf{45.93} & \textbf{47.10} & \textbf{64.91} \\
    \bottomrule
    \end{tabular}%
    \caption{Performance metrics for different models and decoding strategies. \textsc{CoCoA} shows improvements over previous methods across all datasets. More details in Appendix~\ref{app:qaDetails}.}
    \label{tab:model-performance}
\end{table}
\begin{table*}[h]
\centering
\scriptsize
\begin{tabular}{lccccccccc}
\toprule
\multirow{2}{*}{\textbf{Decoding}} & \multicolumn{3}{c}{\textbf{CNN-DM}} & \multicolumn{3}{c}{\textbf{XSum}} & \multicolumn{3}{c}{\textbf{TofuEval (AlignScore)}} \\
\cmidrule(lr){2-4} \cmidrule(lr){5-7} \cmidrule(lr){8-10}
 & ROUGE-L & BERT-P & AlignScore & ROUGE-L & BERT-P & AlignScore & Overall & Main & Marginal \\
\midrule
Greedy     & 24.93 & 95.41 & 91.44 & 14.36 & 94.05 & 85.28 & 76.66 & 81.64 & 61.19 \\
CAD        & 24.76 & 94.45 & 91.01 & 14.59 & 93.65 & 84.34 & 83.23 & 87.26 & 73.58 \\
COIECD     & 23.47 & 92.06 & 85.49 & 13.65 & 91.04 & 73.81 & 60.86 & 68.06 & 58.31 \\
\textsc{AdaCAD}  & 25.42 & 94.91 & 94.97 & 14.91 & 94.29 & 85.81 & 85.07 & 88.06 & \textbf{75.79} \\
\textsc{CoCoA}   & \textbf{25.68} & \textbf{95.42} & \textbf{95.70} & \textbf{15.06} & \textbf{94.60} & \textbf{87.94} & \textbf{86.32} & \textbf{89.14} & 75.51 \\
\bottomrule
\end{tabular}%
\caption{Summarization performance on CNN-DM, XSum and TofuEval with our best performing model (Llama3-70B). \textsc{CoCoA} consistently delivers the highest alignment (AlignScore) and strong ROUGE-L and BERT-P results, outperforming both contrastive and adaptive baselines. Full results across all models are in Table \ref{tab:summarization} in Appendix~\ref{app:sumDetails}.}
\label{tab:summ_llama3_70b}
\end{table*}

\begin{table*}[ht]
    \centering
    \scriptsize
    \tabcolsep3pt
    \begin{tabular}{llccccccccccc}
    \toprule
    \multirow{2}{*}{Model} & \multirow{2}{*}{Decoding} & \multicolumn{2}{c}{CLAPNQ} & \multicolumn{2}{c}{ExpertQA} & \multicolumn{2}{c}{HAGRID} & \multicolumn{2}{c}{ELI5-WebGPT} & \multicolumn{2}{c}{QuoteSum} & Avg. Faith. \\ \cmidrule(lr){3-4} \cmidrule(lr){5-6} \cmidrule(lr){7-8} \cmidrule(lr){9-10} \cmidrule(lr){11-12}
    & & ROUGE-L & Faith & ROUGE-L & Faith & ROUGE-L & Faith & ROUGE-L & Faith & SEMQA & Faith \\
    \midrule
    \textbf{GPT-4o} & Greedy   & 40.53 & 91.81 & 46.34 & 69.48 & 57.76 & 90.86 & 59.04 & 81.00 & 42.56 & 78.51 & 82.33 \\
    \midrule
    \textbf{GPT-4o-mini} & Greedy   & 37.72 & 90.35 & 45.30 & 66.53 & 54.87 & 87.94 & 56.09 & 81.89 & 40.74 & 78.16 & 80.97 \\
    \midrule
    \textbf{Llama-3.1-70B-Instruct} & Greedy   & 39.44 & 88.64 & 43.02 & 69.35 & 49.21 & 79.08 & 51.66 & 74.87 & 41.24 & 67.42 & 75.87 \\
                                     & CAD      & 38.56 & 89.75 & 42.55 & 70.19 & 48.15 & 80.32 & 50.24 & 75.25 & 40.85 & 68.56 & 76.41 \\
                                     & \textsc{AdaCAD}   & 41.00 & 91.32 & 45.03 & 71.12 & 50.04 & 81.57 & 52.12 & 77.11 & 43.06 & 70.23 & 78.27 \\
                                     & \textsc{CoCoA}     & \textbf{42.15} & \textbf{92.45} & \textbf{46.10} & \textbf{72.40} & \textbf{52.07} & \textbf{82.20} & \textbf{54.22} & \textbf{78.95} & \textbf{44.50} & \textbf{71.44} & \textbf{79.49} \\
    \midrule
    \textbf{Mistral-NeMo-12B-Instruct} & Greedy   & 35.28 & 78.71 & 42.76 & 54.19 & 53.05 & 80.16 & 53.84 & 65.06 & 39.50 & 69.85 & 69.59 \\
                                       & CAD      & 34.23 & 79.20 & 41.81 & 55.42 & 51.78 & 79.55 & 52.90 & 66.10 & 38.90 & 69.30 & 69.91 \\
                                       & \textsc{AdaCAD}   & 37.50 & 81.89 & 44.10 & 58.71 & 55.44 & 82.22 & 56.76 & 69.01 & 40.90 & 72.19 & 72.40 \\
                                       & \textsc{CoCoA}     & \textbf{39.75} & \textbf{84.12} & \textbf{46.27} & \textbf{61.20} & \textbf{58.56} & \textbf{83.30} & \textbf{59.89} & \textbf{71.42} & \textbf{42.57} & \textbf{74.55} & \textbf{74.92} \\
    \midrule
    \textbf{Llama-3.1-8B-Instruct} & Greedy   & 17.14 & 58.47 & 31.67 & 51.22 & 16.47 & 55.80 & 47.11 & 55.74 & 25.96 & 41.70 & 52.59 \\
                                     & CAD      & 16.23 & 60.24 & 30.58 & 53.47 & 15.89 & 58.20 & 46.15 & 57.35 & 24.30 & 42.30 & 54.71 \\
                                     & \textsc{AdaCAD}   & 18.43 & 62.37 & 32.87 & 54.76 & 17.12 & 59.90 & 48.22 & 58.71 & 26.80 & 43.57 & 55.86 \\
                                     & \textsc{CoCoA}     & \textbf{19.50} & \textbf{64.18} & \textbf{34.98} & \textbf{57.24} & \textbf{18.40} & \textbf{61.50} & \textbf{50.33} & \textbf{60.24} & \textbf{28.22} & \textbf{45.53} & \textbf{57.74} \\
    \bottomrule
    \end{tabular}%
    \caption{Performance on LFQA datasets showing ROUGE-L (RL) and faithfulness (Faith) scores for Greedy, CAD, \textsc{AdaCAD}, and \textsc{CoCoA} methods. For QuoteSum, we report SEMQA rather than (RL). \textbf{\textsc{CoCoA}} consistently outperforming others on average factuality. More results in Appendix~\ref{app:lfqaDetails}.}
    \label{tab:lfqa-performance-instruct}
\end{table*}

%
\paragraph{QA Tasks.} From Table~\ref{tab:model-performance}, we observe that \textbf{\textsc{CoCoA}} consistently outperforms all baselines (greedy decoding, CAD, COIECD, and the strong \textsc{AdaCAD}) across all QA datasets and model scales. On \textbf{Llama3-70B}, \textsc{CoCoA} yields an average absolute gain of \textbf{11.37 pts} over greedy decoding and \textbf{9.19 pts} over \textsc{AdaCAD}, reflecting its robustness across both high- and low-conflict scenarios.

While CAD often suffers from degraded performance in low-conflict contexts, e.g., an average drop of \textbf{11.3 pts} across tasks on Llama3-8B, \textsc{CoCoA} maintains high accuracy even on mixed or minimal-conflict datasets such as NQ, TriviaQA, and PopQA. Notably, \textsc{CoCoA} surpasses \textsc{AdaCAD} by \textbf{7.9 pts} on \textbf{TriviaQA} and by a substantial \textbf{12.42 pts} on \textbf{PopQA} with Llama3-8B.

On \textbf{NQ-SWAP}, a high-conflict dataset, \textsc{CoCoA} demonstrates its ability to dynamically leverage context-sensitive contrast, achieving \textbf{88.32} with Llama3-70B, well above \textsc{AdaCAD} (\textbf{70.07}) and greedy decoding (\textbf{55.74}). Similarly, on the semi-structured \textbf{TabMWP} dataset, \textsc{CoCoA} reaches \textbf{78.90} with Llama3-70B, outperforming \textsc{AdaCAD} by \textbf{11.8} pts, indicating strong generalization beyond standard QA formats. Across all models, \textsc{CoCoA} consistently achieves the highest average accuracy. These results confirm \textsc{CoCoA}’s effectiveness in balancing contextual information and model confidence across diverse QA conditions.
\paragraph{Summarization Tasks.}
Table~\ref{tab:summ_llama3_70b} evaluates \textsc{CoCoA} on long-form summarization benchmarks (CNN-DM, XSum, and TofuEval) demonstrating consistent improvements over all baselines, including \textsc{AdaCAD}. \textsc{CoCoA} achieves the highest scores across all metrics, highlighting its ability to enhance both surface-level fluency and deeper factual consistency. On \textbf{TofuEval}, a benchmark targeting factual alignment under topic shifts, \textsc{CoCoA} attains the highest AlignScore of \textbf{86.32}, surpassing greedy decoding by \textbf{9.66 pts}, CAD by \textbf{3.09 pts}, COIECD by \textbf{25.46 pts}, and \textsc{AdaCAD} by \textbf{1.25 pts}. Notably, on the more challenging \textit{marginal-topic} subset, \textsc{CoCoA} scores \textbf{75.51}, outperforming all other methods (except \textsc{AdaCAD}), demonstrating its strength in preserving factuality even in less salient content regions where prior methods often drift or hallucinate.

On \textbf{CNN-DM}, \textsc{CoCoA} leads with a ROUGE-L of \textbf{25.68}, BERT-P of \textbf{95.42}, and AlignScore of \textbf{95.70}, showing notable improvements over both COIECD and \textsc{AdaCAD}. On the more abstractive \textbf{XSum}, it yields modest gains in ROUGE-L (\textbf{15.06}) and BERT-P (\textbf{94.60}), but a more pronounced improvement in AlignScore (\textbf{87.94}, +2.13 pts over \textsc{AdaCAD}), indicating stronger factual grounding under abstraction. Together, these results affirm \textsc{CoCoA}’s advantage in generating context-faithful, coherent summaries, especially under factuality-sensitive or out-of-distribution prompts where traditional contrastive methods struggle.
\paragraph{LFQA Tasks.}
Table~\ref{tab:lfqa-performance-instruct} presents \textsc{CoCoA}’s performance across five long-form QA datasets, comparing it with two closed-source baselines (GPT-4o and GPT-4o-mini (both under greedy decoding)) and three open models (Llama-3.1-70B-Inst., Mistral-NeMo-12B-Inst., and Llama-3.1-8B-Inst.) using CAD, AdaCAD, and \textsc{CoCoA}. Despite operating on open models, \textsc{CoCoA} achieves results competitive with, and in some cases surpassing, state-of-the-art closed systems. For instance, on Llama-3.1-70B with CLAPNQ, \textsc{CoCoA} attains a ROUGE-L of \textbf{42.15}, outperforming GPT-4o (40.53), and reaches a FaithScore of \textbf{92.45}, slightly above GPT-4o’s 91.81. Against GPT-4o-mini on CLAPNQ, \textsc{CoCoA} shows even clearer gains (+4.43 ROUGE-L, +2.10 FaithScore), highlighting its ability to elevate smaller open models to the performance tier of much larger proprietary systems. Appendix~\ref{app:caseStudy} shows a qualitative example of improved response generation by \textsc{CoCoA}.

Compared to previous open decoding approaches, \textsc{CoCoA} consistently improves both relevance and factuality. On Llama-3.1-70B, it raises the average ROUGE-L across datasets from 47.05 (AdaCAD) to \textbf{48.64} and the FaithScore from 78.27 to \textbf{79.49}. Mistral-NeMo-12B exhibits similar trends: \textsc{CoCoA} improves ROUGE-L by +2.67 pts and FaithScore by +2.11 pts over AdaCAD on average all datasets. Even with the smaller Llama-3.1-8B model, \textsc{CoCoA} delivers gains of +1.64 pts ROUGE-L and +1.88 pts FaithScore. These improvements are consistent across all datasets (CLAPNQ, ExpertQA, HAGRID, ELI5-WebGPT, and QuoteSum), reinforcing \textsc{CoCoA}’s effectiveness in balancing parametric knowledge and retrieved context for grounded, coherent generation. 

Although, \textsc{CoCoA} based decoding from Llama-3.1-70B-Instruct beats all other open source models, it still performs slightly worse on average compared to GPT-40-mini and GPT-4o.

\begin{table}[!t]
\centering
\scriptsize
\begin{tabular}{@{}lcc@{}}
\toprule
\textbf{Model Variant}                 & \textbf{NQ} & \textbf{NQ-SWAP} \\
\midrule
Full \textsc{CoCoA}                            & \textbf{49.30}          & \textbf{79.15}               \\
w/o Rényi w/ KL Divergence                  &     48.20      &   76.69            \\
w/o Rényi Divergence                  & 47.10          & 70.80               \\
w/o Entropy Gap                       & 46.85          & 68.90               \\
w/o Contextual peakedness              & 45.60          & 65.20               \\
w/o Adaptive Gating ($\lambda$=0.5) & 44.75          & 62.50               \\
Greedy Decoding (Baseline)            & 44.63          & 47.81 \\
\bottomrule
\end{tabular}%
\caption{Exact Match (EM) accuracy of Llama3-8B on Natural Questions (NQ) and NQ-SWAP under different ablations of the \textsc{CoCoA} decoding framework. Each variant removes or alters a core component to assess its individual contribution to performance.}
\label{tab:ablation}
\end{table}

\begin{table}[!b]
\centering
\scriptsize
\begin{tabular}{lccc}
\toprule
\textbf{Decoding} & \textbf{NQ-SWAP} & \textbf{NQ-SYNTH} & \textbf{Overall} \\
\midrule
Greedy   & 51.60 & 88.20 & 69.90 \\
CAD      & 79.60 & 64.00 & 71.80 \\
COIECD   & 50.80 & 83.60 & 67.20 \\
\textsc{AdaCAD}   & 62.80 & 86.40 & 74.60 \\
\textsc{CoCoA}     & \textbf{80.80} & \textbf{86.60} & \textbf{83.70}\\
\bottomrule
\end{tabular}%
\caption{Accuracy on conflicting data (NQ-SWAP) and non-conflicting data (NQ-SYNTH) with Llama3-70B.}
\label{tab:conflict}
\end{table}

\begin{table*}[ht]
    \centering
    \tiny
    \begin{tabular}{|p{0.9\textwidth}|}
    \hline
    \textbf{Context}: \textcolor{hallucination}{Tom Hanks played multiple roles in The Polar Express, providing motion-capture performances for the Hero Boy, the Hero Boy’s father, the Conductor, the Hobo, Santa Claus, and the Narrator.} \textcolor{blue}{Daryl Sabara} as the Hero Boy. ($\ldots$). The voice of the Hero Boy ($\ldots$) \textcolor{hallucination}{Dylan Cash} as Boy on Train (voice) ($\ldots$) \textcolor{LimeGreen}{Dante Pastula played the Little Boy}. A group of ($\ldots$) motion-capture.\\
\textbf{Question}: Who played the little boy in polar express ?\\
\textbf{Gold Answer}: \textcolor{LimeGreen}{Dante Pastula} | \textbf{Parametric Knowledge (Llama3-70B)}: \textcolor{blue}{Daryl Sabara}.\\
\textbf{Greedy}: \textcolor{hallucination}{Dylan Cash} | \textbf{AdaCAD}: \textcolor{hallucination}{Tom Hanks} | \textbf{CoCoA}: \textcolor{LimeGreen}{Dante Pastula}\\
\hline
\textbf{Context (Meeting Transcripts)}: \\
Speaker 0: Agenda Item One: Council Bill 119945, concerning the Civil Service Commissions. This bill amends election procedures for the employee-elected members of both the Civil Service Commission and the Public Safety Civil Service Commission, and includes temporary provisions for the 2020 elections. It amends Sections 4.04.250 and 4.08.040 of the Seattle Municipal Code, and ratifies and confirms certain prior actions. Thank you, Madam Chair. I move to pass Council Bill 119945. Is there a second? \\
Speaker 1: Second. \\
($\ldots$) \\
\textcolor{LimeGreen}{Both commissions are holding elections this year, but due to COVID-19-related delaysIn addition, the bill updates outdated language in the Seattle Municipal Code and empowers the City Clerk to modernize election procedures through future} \\
Speaker 0: Nine in favor, none opposed. The bill passes. ($\ldots$)  \\
\textbf{Greedy Decoding}: The bill amends election procedures for the Civil Service Commission and Public Safety Civil Service Commission, allowing the city clerk to hold elections in December instead of November, and permitting electronic ballots. The changes are temporary and only apply to the 2020 elections.\\
\textbf{AdaCAD}: This bill amends Seattle Municipal Code (SMC) 4.04.250 and 4.08.040 to allow the City Clerk to hold elections for the Civil Service Commission and Public Safety Civil Service Commission in December 2020 rather than November 2020. The bill also allows for electronic ballots rather than only physical ballots, though the option of paper ballots will still remain, and the schedule changes only for 2020. \textcolor{hallucination}{And these elements are only included in the ordinance, not the}\\
\textbf{CoCoA}: The bill updates election procedures for employee-elected members of the Civil Service and Public Safety Civil Service Commissions, allowing electronic ballots and rescheduling 2020 elections to December due to \textcolor{LimeGreen}{COVID-19, while also removing outdated code language and enabling future modernization} by the City Clerk.\\
\hline
    \end{tabular}
    \caption{\textit{Qualitative examples showcasing \textbf{\textsc{CoCoA}}'s improved factual grounding and conflict resolution.} In the NQ example (top), \textsc{CoCoA} predicts the correct answer (\textcolor{LimeGreen}{green}) by aligning contextual and parametric signals (\textcolor{blue}{blue}), unlike greedy and \textsc{AdaCAD} (\textcolor{hallucination}{red}). In the TofuEval case (bottom), \textsc{CoCoA} avoids hallucinations (\textcolor{hallucination}{red}) and remains faithful to the source (\textcolor{LimeGreen}{green}).}
    \label{tab:qualitative}
\end{table*}

\subsection{Ablation Study}
As shown in Tab.~\ref{tab:ablation}, (1) Rényi Divergence is crucial for detecting nuanced conflicts between the model's knowledge and the context. (2) Entropy Gap provides valuable insights into the confidence level introduced by the context. (3) Contextual peakedness effectively strengthens the model's reliance on highly confident contextual information. (4) Adaptive Gating ensures a balanced integration of the model's prior knowledge and the context, adapting to the specific needs of each decoding step. More details in Appendix \ref{sec:ablation}.

\begin{table}[!b]
\centering
\scriptsize
\begin{tabular}{lccc}
\toprule
\textbf{Decoding} & $\rho$ (NQ-SWAP) & $\rho$ (NQ-SYNTH) & $|\Delta \rho|$ \\
\midrule
CAD     & 0.56 & 0.57 & 0.01 \\
\textsc{AdaCAD}  & 0.86 & 0.94 & 0.08 \\
\textsc{CoCoA}     & 0.74 & 0.95 & \textbf{0.21} \\
\bottomrule
\end{tabular}%
\caption{Spearman rank-order correlation coefficient between original and adjusted output distributions for \textsc{CAD}, \textsc{AdaCAD} and \textsc{CoCoA} on NQ-SWAP and NQ-SYNTH. The difference $|\Delta \rho|$ measures the sensitivity of a decoding method to the degree of conflict (higher is better).}
\label{tab:spearman}
\end{table}


\section{Further Analysis}
\subsection{High vs. Low Conflict Instances}
\paragraph{Setup.} To evaluate decoding strategies across varying levels of conflict between context and a model's internal knowledge, we employed \textbf{NQ-SYNTH} (non-conflicting) and \textbf{NQ-SWAP} (highly conflicting) datasets. These datasets (totaling to 500 samples), established in prior work~\citep{wang-etal-2025-adacad}, isolate conflict as the primary challenge, allowing us to assess whether decoding methods like \textsc{CoCoA} can generalize effectively across these distinct regimes.
\paragraph{Result.} As shown in Table~\ref{tab:conflict}, \textsc{CoCoA} demonstrates a significant advantage over prior decoding methods on both high-conflict (NQ-SWAP) and low-conflict (NQ-SYNTH) examples. Unlike \textsc{CAD}, which showed a notable performance drop of 24.2 pts on NQ-SYNTH compared to greedy decoding, \textsc{CoCoA} maintained near-optimal accuracy on non-conflicting data (86.6\%) relative to greedy (88.2\%). Crucially, \textsc{CoCoA} achieved the highest accuracy on NQ-SWAP (80.8\%), surpassing \textsc{AdaCAD} by 18.0 pts, \textsc{COIECD} by 30.0 pts, and greedy decoding by 29.2 pts. This indicates that \textsc{CoCoA} effectively preserves fidelity in low-conflict scenarios without unduly penalizing agreement, a common issue with methods like \textsc{CAD} and \textsc{COIECD}. While \textsc{AdaCAD} aimed for adaptability across conflict levels, \textsc{CoCoA} leverages divergence-based normalization and entropy-aware amplification for a more granular, token-level calibration. This fine-tuning is particularly beneficial in long-context QA with mixed evidence. Consequently, \textsc{CoCoA} offers a robust decoding policy that generalizes effectively across both adversarial and naturally-aligned inputs, yielding an absolute overall accuracy gain of 9.1 pts over \textsc{AdaCAD}.

\subsection{JSD does not adequately address conflict}
While \textsc{AdaCAD} improves over CAD via a JSD-based gating mechanism, it remains limited in detecting fine-grained or evolving conflicts between model priors and external context. To assess conflict sensitivity, we compute the Spearman correlation $\rho$ between each method's output distribution and that of greedy decoding across NQ-SWAP (high conflict) and NQ-SYNTH (low conflict). An ideal context-aware decoder should show low $\rho$ in high-conflict and high $\rho$ in low-conflict settings. As shown in Tab.~\ref{tab:spearman}, \textsc{AdaCAD} exhibits a narrow correlation gap ($|\Delta \rho| = 0.08$), indicating limited responsiveness to conflict.

\begin{figure}[!t]
    \centering
    \includegraphics[width=\linewidth]{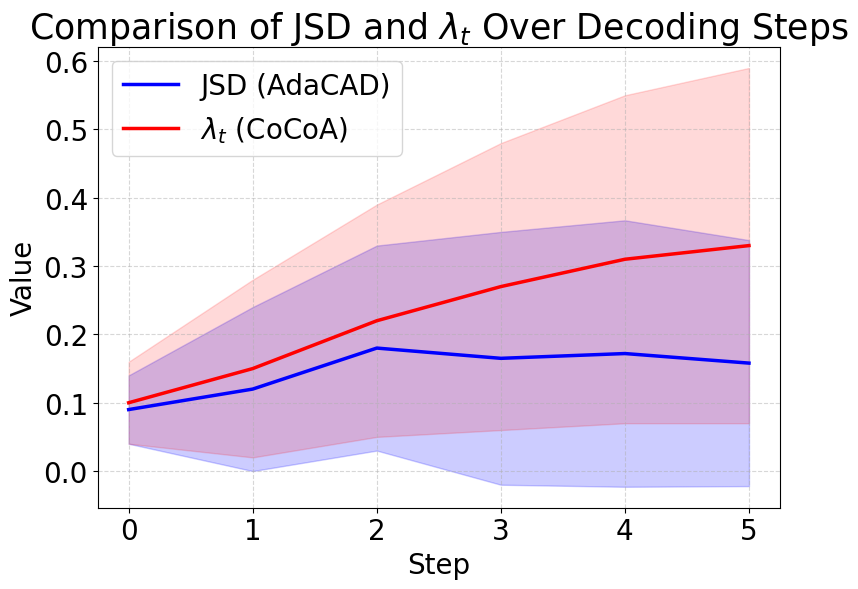}
    \caption{Comparison of JSD values (used in \textsc{AdaCAD}) and $\lambda_t$ values (used in \textsc{CoCoA}) over the first 5 decoding steps ($t$) using LLaMA3-70B on TofuEval. } 
    \label{fig:conflict-trace}
\end{figure}

\textsc{CoCoA} achieves a substantially larger gap ($|\Delta \rho| = 0.21$), driven by three enhancements: (1) Rényi divergence, which better captures low-probability, contextually relevant alternatives; (2) an entropy gap to model uncertainty shifts; and (3) contextual peakedness to emphasize strong contextual cues. Fig.~\ref{fig:conflict-trace} illustrates \textsc{CoCoA}'s more dynamic conflict signal $\lambda_t$, which evolves with context and shows greater sensitivity than the plateaued JSD trace in \textsc{AdaCAD}. This enables \textsc{CoCoA} to adaptively modulate generation in response to context-model divergence.
\subsection{Qualitative Analysis and Case Study}
We provide representative examples from NQ and TofuEval in Tab.~\ref{tab:qualitative} to illustrate \textsc{CoCoA}’s ability to handle both QA and summarization tasks under knowledge conflict. In the NQ example, greedy decoding and \textsc{AdaCAD} both fail due to incorrect parametric priors, producing ``Dylan Cash'' and ``Tom Hanks'' respectively, while \textsc{CoCoA} successfully recovers the correct answer ``Dante Pastula'' by down-weighting conflicting parametric knowledge and emphasizing grounded context. In the summarization case, \textsc{CoCoA} better captures the COVID-related election delays and modernization efforts, faithfully aligning with the highlighted source sentences. In contrast, \textsc{AdaCAD} introduces unrelated constraints and Greedy omits key updates. These examples underscore \textsc{CoCoA}’s advantage in adaptively reconciling model knowledge and context evidence, especially under fine-grained or subtle conflicts.

\section{Discussion and Conclusion}
\textsc{CoCoA} advances decoding by finely detecting and adapting to varying degrees of knowledge conflict between model memory and context. Its entropy-based, token-level conflict measures allow dynamic blending of parametric and contextual signals, avoiding the pitfalls of static or coarse conflict handling. This results in more faithful and accurate generation across QA, summarization, and long-form tasks, without degrading performance when conflict is low. \textsc{CoCoA}'s principled approach offers a robust, flexible framework for improving context-aware language generation and sets a strong foundation for future dynamic decoding research.
\section*{Limitations}
\textsc{CoCoA} relies on fine-grained access to a model’s token-level probability distributions, with and without retrieved context, to compute Rényi divergences, entropy gaps, and contextual peakedness. This requirement poses a barrier when working with fully black-box APIs (e.g., GPT-4), which typically expose only sampled text rather than the underlying logits or softmax scores. Developing techniques to approximate or infer these distributions without direct logit access would be a valuable extension, enabling broader applicability to proprietary or mobile-only LLM services.

Our current study is also confined to English-language benchmarks and a handful of widely-used open-source models. As large-scale models emerge in other languages and specialized domains (e.g., legal, medical, or code generation), it will be important to validate \textsc{CoCoA}’s conflict-detection and adaptation mechanisms under different linguistic characteristics and domain-specific knowledge structures. Additionally, investigating how model size, instruction-tuning, and alignment procedures interact with \textsc{CoCoA}’s entropy- and margin-based signals could reveal further refinements or simplifications.

Finally, while we do not identify any direct ethical or safety concerns with contrastive, context-aware decoding itself, future work should examine potential biases in context selection (retrieval quality) and ensure that \textsc{CoCoA}’s stronger reliance on external knowledge does not inadvertently amplify misleading or harmful content.
\bibliography{references}

\noindent{\Large \textbf{Overview of Appendices}}
\begin{itemize}
    \item Appendix~\ref{sec:datasets_appendix}: Dataset Details.
    \item Appendix~\ref{sec:prompts}: Prompts
    \item Appendix~\ref{app:qaDetails}: Performance Comparison of Instruction-Tuned Models on QA Benchmarks.
    \item Appendix~\ref{app:sumDetails}: Performance Comparison on Summarization Tasks
    \item Appendix~\ref{app:lfqaDetails}: Performance Comparison on Long-Form QA (LFQA) Datasets
    \item Appendix~\ref{sec:ablation}: Ablation Study
    \item Appendix~\ref{app:caseStudy}: Case Study
    \item Appendix~\ref{app:sensitivity}: Sensitivity Evaluation
    \item Appendix~\ref{sec:latency}: Latency Considerations
\end{itemize}

\appendix
\section{Dataset Details}
\label{sec:datasets_appendix}
We evaluated our approach on a diverse set of  question answering (QA) and summarization datasets, adhering to the experimental setup established by AdaCAD ~\cite{wang-etal-2025-adacad} to enable apples-to-apples comparison. Additionally, we  benchmarked on long-form question answering (LFQA) datasets to evaluate faithfulness, defined as the extent to which the model’s response is factually grounded in the provided context document. We also present one example from each dataset, as detailed in Table~\ref{tab:dataset-prompts}. For the synthetically generated QA datasets NQ-SWAP and NQ-SYNTH, we provide examples in Table \ref{tab:nq-swap-synth-example}.
\begin{table}[t]
\centering
\scriptsize
\begin{tabular}{p{0.97\columnwidth}}
\toprule
\textbf{Question:} \\
Who wrote we're going on a bear hunt ? \\
\midrule
\multicolumn{1}{c}{\textsc{\textbf{Natural Question}}} \\
\textbf{Original Context:} \\
We 're Going on a Bear Hunt is a 1989 children 's picture book written by \textcolor{cyan}{Michael Rosen} and illustrated by Helen Oxenbury .\ldots \\
\textbf{Original Answer:} \textcolor{cyan}{Michael Rosen} \\
\midrule
\multicolumn{1}{c}{\textsc{\textbf{NQ-Swap}}} \\
\textbf{Substitute Context:} \\
We 're Going on a Bear Hunt is a 1989 children 's picture book written by \textcolor{orange}{Robert Hooke} and illustrated by Helen Oxenbury  \ldots \\
\textbf{Substitute Answer:} \textcolor{orange}{Robert Hooke} \\
\midrule
\multicolumn{1}{c}{\textsc{\textbf{NQ-Synth}}} \\
\textbf{Substitute Context:} \\
\textbf{Original Context:} \\
We 're Going on a Bear Hunt is a 1989 children 's picture book written by \textcolor{purple}{Brian Urlacher} and illustrated by Helen Oxenbury .\ldots \\
\textbf{Substitute Answer (generated from LLM):} \textcolor{purple}{Brian Urlacher} \\
\bottomrule
\end{tabular}
\caption{
Example from \textsc{NQ-Swap} and \textsc{NQ-Synth}. A \textcolor{orange}{\textit{substitute example}} for \textsc{NQ-Swap} is made from the \textcolor{cyan}{\textit{original example}} by replacing the original answer, \textcolor{cyan}{Michael Rosen}, with a similar but conflicting answer, i.e., \textcolor{orange}{Robert Hooke}. A \textcolor{purple}{\textit{substitute example}} for \textsc{NQ-Synth} is made from the \textcolor{cyan}{\textit{original example}} by replacing the original answer, \textcolor{cyan}{Michael Rosen}, with one \textcolor{purple}{generated by Llama3-70B without context}, i.e., \textcolor{purple}{Brian Urlacher}.
}
\label{tab:nq-swap-synth-example}
\end{table}
\subsection{Question Answering Datasets}
\paragraph{Natural Questions (NQ)}~\cite{kwiatkowski2019natural} A large-scale QA dataset comprising real anonymized queries from Google Search, each paired with a Wikipedia page. Annotators provide long and short answers based on the content of the page. Following AdaCAD~\cite{wang-etal-2025-adacad}, We utilize a subset of 3,231 validation instances featuring short answers .

\paragraph{NQ-SWAP}~\cite{longpre2021entity} A synthetic variant of NQ designed to introduce knowledge conflicts by replacing named entities in the context with alternate entities, challenging models to handle conflicting information. Following AdaCAD~\cite{wang-etal-2025-adacad}, this dataset contains 4,000 instances derived from NQ.

\paragraph{TriviaQA}~\cite{joshi2017triviaqa} A challenging reading comprehension dataset with over 650K question-answer-evidence triples. Questions are authored by trivia enthusiasts and paired with evidence documents gathered independently. Following AdaCAD~\cite{wang-etal-2025-adacad} we perform evaluation on their sampled 1,000 instances from the Wikipedia domain for evaluation.

\paragraph{PopQA}~\cite{mallen2023popqa} An open-domain QA dataset consisting of 14,000 question-answer pairs focused on long-tail entities. Each instance includes fine-grained Wikidata entity IDs and relationship type information. We perform benchmarking on the selected 1,600 instances as used in AdaCAD~\cite{wang-etal-2025-adacad} where the context contains the gold answer .

\paragraph{HotpotQA}~\cite{yang2018hotpotqa} A multi-hop QA dataset requiring reasoning over multiple supporting documents. It includes 113,000 question-answer pairs with sentence-level supporting facts to facilitate explainable QA systems. We use the full development set comprising 7,405 instances.

\paragraph{TabMWP}~\cite{lu2023tabmwp} A dataset containing 38,431 tabular math word problems that require mathematical reasoning on both textual and tabular data. Each question is aligned with a tabular context presented as an image, semi-structured text, or a structured table. We utilize the ``test1k'' subset, which includes 1,000 instances .

\subsection{Summarization Datasets}
\paragraph{CNN/DailyMail (CNN-DM)}~\cite{see2017get} An English-language dataset containing over 300,000 news articles from CNN and the Daily Mail, paired with multi-sentence summaries. It supports both extractive and abstractive summarization tasks. We use the same sampled (as obtained from AdaCAD~\cite{wang-etal-2025-adacad} authors) set of 500 examples from the test set.

\paragraph{XSum} The Extreme Summarization dataset ~\cite{narayan2018don} comprises 226,711 BBC news articles, each accompanied by a one-sentence summary. The dataset covers a wide range of topics and is designed for evaluating abstractive summarization systems. We perform benchmarking on 500 instances as obtained from AdaCAD~\cite{wang-etal-2025-adacad} authors.

\paragraph{TofuEval}~\cite{tang2024tofu} A benchmark for evaluating the factual consistency and topic relevance of summaries, particularly in dialogue or meeting transcription scenarios. It includes 50 documents each from MediaSum and MeetingBank datasets, with three topics per document, resulting in 300 topic-focused summaries.

\subsection{Long-Form Question Answering (LFQA) Datasets}

\paragraph{ELI5-WebGPT}~\cite{nakano2021webgpt} ELI5-WebGPT is a dataset designed for evaluating long-form question answering (LFQA) systems. It comprises 271 questions sourced from the ``Explain Like I’m Five'' (ELI5) subreddit, as released by WebGPT (Nakano et al., 2021). Each question is paired with human-labeled answers and ``gold'' documents—relevant and informative passages collected by human annotators using commercial search engines like Bing. These documents serve as high-quality evidence to assess the performance of retrieval-augmented models in generating accurate and informative responses.

\paragraph{ExpertQA}~\cite{malaviya2023expertqa} ExpertQA consists of 528 open-ended, information-seeking questions across 32 topics, each paired with relevant documents and expert-verified answers.

\paragraph{HAGRID}~\cite{chen2023hagrid} HAGRID (Human-in-the-loop Attributable Generative Retrieval for Information-seeking Dataset) is a dataset designed for generative information-seeking tasks. It comprises questions paired with manually labeled relevant documents and answers generated by GPT-3.5. These answers are formatted with in-context citations referencing the supporting documents. Human annotators evaluate each answer based on two criteria: informativeness (whether the answer provides a direct response to the question) and attributability (whether the answer's claims are supported by the cited documents). We selected 496 samples where the answers are considered both informative and well-grounded.

\paragraph{CLAPNQ}~\cite{rosenthal2025clapnq}.CLAPNQ (Cohesive Long-form Answers from Passages in Natural Questions) is a benchmark dataset designed to evaluate Retrieval-Augmented Generation (RAG) systems. It comprises 4,946 real web search queries sourced from the Natural Questions dataset, each paired with a single gold passage from Wikipedia and a concise, cohesive long-form answer. These answers are typically 2–3 sentences long and are crafted by integrating non-contiguous parts of the passage to ensure fluency and factual grounding. CLAPNQ supports comprehensive evaluation of RAG systems across retrieval, generation, and full pipeline tasks. For our evaluation, we selected the 300 answerable questions from the CLAPNQ test set, which consists of 600 questions in total (300 answerable and 300 unanswerable). This subset allows for focused assessment of model performance on questions with available grounded answers.

\paragraph{QuoteSum}~\cite{schuster-etal-2024-semqa} QuoteSum is a semi-extractive long-form question answering (LFQA) dataset designed to assess models' abilities to generate grounded, multi-source answers. Each question is accompanied by relevant documents and human-written answers that explicitly incorporate extracted spans from multiple sources. These answers blend verbatim quotes with connective text to form cohesive, well-grounded responses. The dataset emphasizes the Semi-Extractive Multi-Source Question Answering (SEMQA) task, challenging models to synthesize information while maintaining precise attributions. The test subset of QuoteSum comprises 1,319 examples, each consisting of a question, a set of relevant documents, and a human-written answer. This subset is used to assess the performance of retrieval-augmented generation (RAG) systems in generating accurate and informative answers multiple sources.
\subsection{Licenses}
The datasets employed in our study are distributed under the following licenses:
\begin{itemize}
  \item \textbf{Natural Questions (NQ)}: Apache License 2.0
  \item \textbf{NQ-SWAP}: MIT License
  \item \textbf{TriviaQA}: Apache License 2.0
  \item \textbf{PopQA}: MIT License
  \item \textbf{HotpotQA}: Apache License 2.0
  \item \textbf{TabMWP}: MIT License
  \item \textbf{CNN/DailyMail (CNN-DM)}: Apache License 2.0
  \item \textbf{XSum}: MIT License
  \item \textbf{TofuEval}: MIT License
  \item \textbf{ELI5}: Creative Commons Attribution-ShareAlike 4.0 International License
  \item \textbf{ExpertQA}: MIT License
  \item \textbf{HaGRID}: Apache License 2.0
  \item \textbf{CLAPNQ}: Apache License 2.0
  \item \textbf{QuoteSum}: Creative Commons Attribution-ShareAlike 4.0 International License
\end{itemize}
The models utilized in our experiments are governed by the following licenses:
\begin{itemize}
  \item \textbf{LLaMA 2}: Custom license available at \url{https://ai.meta.com/llama/license/}
  \item \textbf{LLaMA 3}: Custom license available at \url{https://www.llama.com/llama3/license/}
  \item \textbf{Mistral}: Apache License 2.0
\end{itemize}

\section{Prompts}
\label{sec:prompts}
\begin{figure*}[!t]
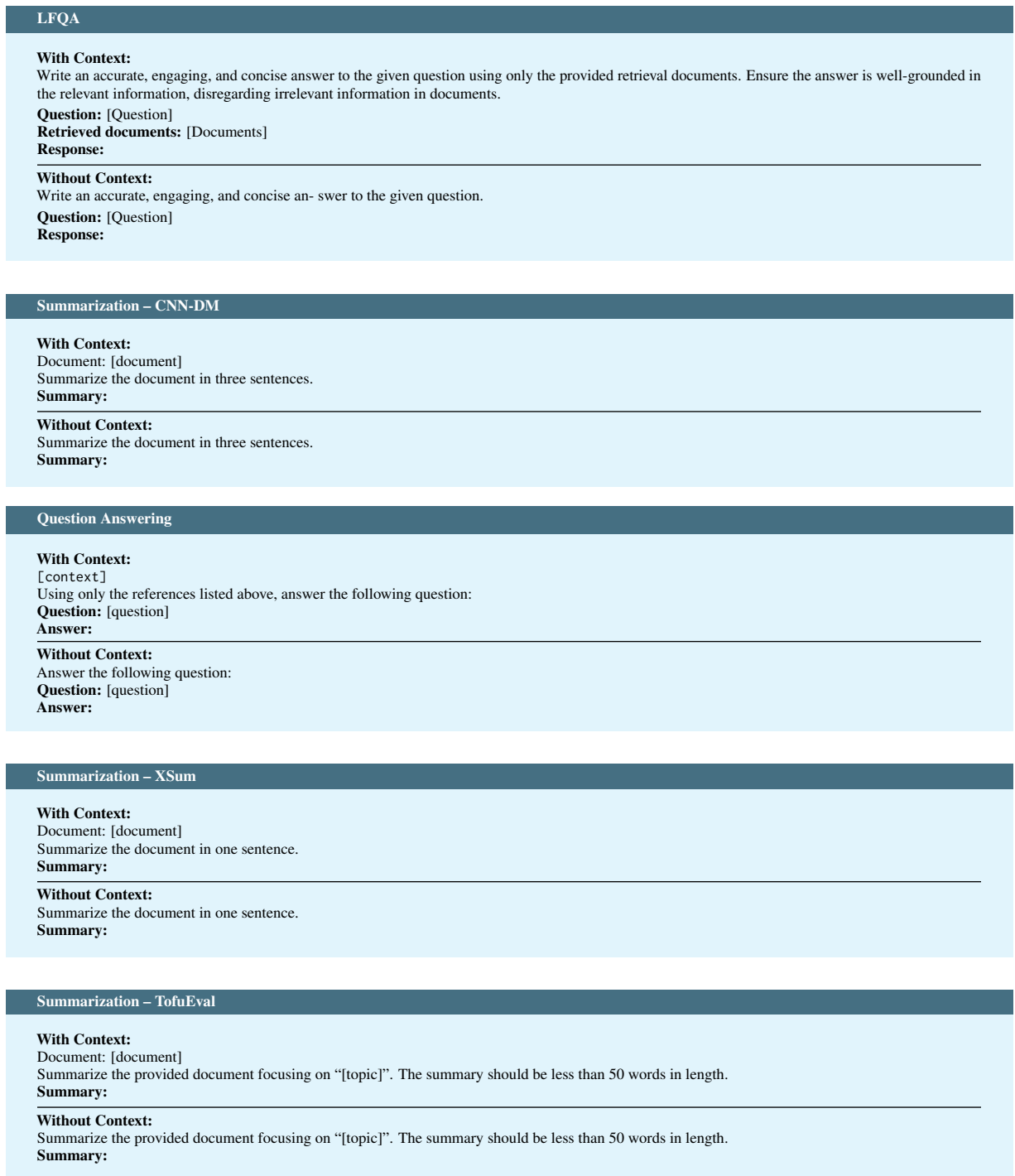

\centering
\scriptsize
\begin{minipage}[t]{\textwidth}
  \begin{tcolorbox}[
    title=LFQA,
    colback=cyan!10, colframe=cyan!40!black,
    fonttitle=\bfseries, sharp corners,
    boxrule=0.8pt,
    leftrule=0pt, rightrule=0pt, toprule=0pt, bottomrule=0pt
  ]
    \textbf{With Context:} \\
    Write an accurate, engaging, and concise answer to the given question using only the provided retrieval documents. Ensure the answer is well-grounded in the relevant information, disregarding irrelevant information in documents.\\[0.5ex]
    \textbf{Question:} [Question]\\
    \textbf{Retrieved documents:} [Documents]\\
    \textbf{Response:}
    \vspace{1ex}
    \hrule height 0.5pt
    \vspace{1ex}
    \textbf{Without Context:} \\
    Write an accurate, engaging, and concise an-
swer to the given question.\\[0.5ex]
    \textbf{Question:} [Question]\\
    \textbf{Response:}
  \end{tcolorbox}

  \vspace{1em}

  \begin{tcolorbox}[
    title=Summarization – CNN-DM,
    colback=cyan!10, colframe=cyan!40!black,
    fonttitle=\bfseries, sharp corners,
    boxrule=0.8pt,
    leftrule=0pt, rightrule=0pt, toprule=0pt, bottomrule=0pt
  ]
    \textbf{With Context:} \\
    Document: [document]\\
    Summarize the document in three sentences.\\
    \textbf{Summary:}
    \vspace{1ex}
    \hrule height 0.5pt
    \vspace{1ex}
    \textbf{Without Context:}\\
    Summarize the document in three sentences.\\
    \textbf{Summary:}
  \end{tcolorbox}
\end{minipage}
 \vspace{1em}
 
\begin{minipage}[t]{\textwidth}
  \begin{tcolorbox}[
    title=Question Answering,
    colback=cyan!10, colframe=cyan!40!black,
    fonttitle=\bfseries, sharp corners,
    boxrule=0.8pt,
    leftrule=0pt, rightrule=0pt, toprule=0pt, bottomrule=0pt
  ]
    \textbf{With Context:} \\
    \texttt{[context]}\\
    Using only the references listed above, answer the following question: \\
    \textbf{Question:} [question]\\
    \textbf{Answer:}
    \vspace{1ex}
    \hrule height 0.5pt
    \vspace{1ex}
    \textbf{Without Context:}\\
    Answer the following question: \\
    \textbf{Question:} [question]\\
    \textbf{Answer:}
  \end{tcolorbox}

  \vspace{1em}

  \begin{tcolorbox}[
    title=Summarization – XSum,
    colback=cyan!10, colframe=cyan!40!black,
    fonttitle=\bfseries, sharp corners,
    boxrule=0.8pt,
    leftrule=0pt, rightrule=0pt, toprule=0pt, bottomrule=0pt
  ]
    \textbf{With Context:} \\
    Document: [document]\\
    Summarize the document in one sentence.\\
    \textbf{Summary:}
    \vspace{1ex}
    \hrule height 0.5pt
    \vspace{1ex}
    \textbf{Without Context:}\\
    Summarize the document in one sentence.\\
    \textbf{Summary:}
  \end{tcolorbox}

  \vspace{1em}

  \begin{tcolorbox}[
    title=Summarization – TofuEval,
    colback=cyan!10, colframe=cyan!40!black,
    fonttitle=\bfseries, sharp corners,
    boxrule=0.8pt,
    leftrule=0pt, rightrule=0pt, toprule=0pt, bottomrule=0pt
  ]
    \textbf{With Context:} \\
    Document: [document]\\
    Summarize the provided document focusing on “[topic]”. The summary should be less than 50 words in length.\\
    \textbf{Summary:}
    \vspace{1ex}
    \hrule height 0.5pt
    \vspace{1ex}
    \textbf{Without Context:}\\
    Summarize the provided document focusing on “[topic]”. The summary should be less than 50 words in length.\\
    \textbf{Summary:}
  \end{tcolorbox}
\end{minipage}
\caption{Prompt templates for QA, summarization and LFQA tasks, with and without context.}
\label{fig:prompt-templates}
\end{figure*}

The prompts used on pre-trained base language model with and without context for QA, LFQA and summarization tasks are given in Figure \ref{fig:prompt-templates}.
\begin{table*}[t]
\centering
\scriptsize
\begin{tabular}{p{0.17\textwidth} | p{0.78\textwidth}}
\toprule
\textbf{Natural Question} &
\textbf{c:} Tom Hanks performed the motion capture for multiple characters including the Hero Boy, the Hero Boy’s father, the Conductor, the Hobo, Santa Claus, and the Narrator. Daryl Sabara provided the voice for the Hero Boy, while Josh Hutcherson contributed additional motion capture for the same character \dots \newline
\textbf{x:} Who played the little boy in polar express? \\
\midrule

\textbf{NQ-SWAP} &
\textbf{c:} The image of the gates in popular culture is a set of large gold , white or wrought - iron gates in the clouds , guarded by Conor Maynard ( the keeper of the `` keys to the kingdom '' ) . Those not fit to enter heaven are denied entrance at the gates , and descend into Hell . In some versions of this imagery , Peter looks up the deceased 's name in a book , before opening the gate  \dots \newline
\textbf{x:} Who do you meet at the gates of heaven ? \\
\midrule

\textbf{TriviaQA} &
\textbf{c:} \dots Colin Baker had been signed up for four years , as the previous actor Peter Davison had left after only three years . Prior to its postponement , season 23 was well advanced with episodes already drafted and in at least one case distributed to cast and production . Alongside `` The Nightmare Fair '' , The Ultimate Evil '' , `` Mission to Magnus '' , `` Yellow Fever and How to Cure It '' , the \dots \newline
\textbf{x:} Which actor was the fifth Doctor Who from 1982-1984, and in that role often wore Edwardian cricket costume? \\
\midrule

\textbf{PopQA} &
\textbf{c:} Charles Towneley Strachey, 4th Baron O'Hagan (born 6 September 1945) is a British Conservative party politician.  \newline
\textbf{x:}  What is Charles Strachey, 4th Baron O'Hagan's occupation? \\
\midrule

\textbf{HotpotQA} &
\textbf{c:} \dots <t> Front Row (software) </t> Front Row is a discontinued media center software application for Apple's Macintosh computers and Apple TV for navigating and viewing video, photos, podcasts, and music from a computer, optical disc, or the Internet through a 10-foot user interface (similar to Windows Media Center and Kodi). The software relies on iTunes and iPhoto and is controlled by an Apple Remote or the keyboard function keys. \dots \newline
\textbf{x:} Aside from the Apple Remote, what other device can control the program Apple Remote was originally designed to interact with? \\
\midrule

\textbf{TabMWP} &
\textbf{c:} Table: blender | \$14.02 CD | \$18.35 computer mouse | \$10.65 CD player | \$21.84 DVD player | \$53.57 radio | \$15.42 \newline
\textbf{x:} Roxanne has \$32.50. Does she have enough to buy a CD and a blender? \\
\midrule

\textbf{CNN-DM} &
\textbf{c:} Article: (CNN)Malala Yousafzai's stellar career has included a Nobel Peace Prize. Last week, she made it into outer space. A NASA astrophysicist has named an asteroid after the teenage education activist from Pakistan, who was gravely wounded by a Pakistani Taliban \dots \newline
\textbf{x:} Summarize the article in three sentences. \textbf{Summary:} \\
\midrule

\textbf{XSum} &
\textbf{c:} The full cost of damage in Newton Stewart, one of the areas worst affected, is still being assessed. Repair work is ongoing in Hawick and many roads in Peeblesshire remain badly affected by standing water. Trains on the west coast mainline face disruption due to damage at the Lamington Viaduct. Many businesses and householders were affected by flooding in Newton Stewart after the River Cree overflowed into the town. \dots \newline
\textbf{x:} Summarize the article in one sentence. \textbf{Summary:} \\
\midrule

\textbf{TofuEval} &
\textbf{c:} Document: DOBBS: Coming up at the top of the hour here on CNN, ``THE SITUATION ROOM'' with Wolf Blitzer. Here's Wolf -- Wolf. WOLF BLITZER, CNN ANCHOR: Thanks very much, Lou. Poker, hookers and the CIA? Police search the home of the man who was the third in charge over at the Central Intelligence Agency \dots \newline
\textbf{x:} Summarize the provided document focusing on ``Poker, Hookers, and the CIA''. The summary should be less than 50 words in length. \textbf{Summary:} \\
\bottomrule
\end{tabular}
\caption{Examples of prompt templates from various QA and summarization datasets. `c:' denotes the context (document, table, or passage), and `x:' denotes the corresponding question or summarization instruction.}
\label{tab:dataset-prompts}
\end{table*}

\section{Performance Comparison of Instruction-Tuned Models on QA Benchmarks}
\label{app:qaDetails}
We evaluate the performance of various instruction-tuned language models on multiple QA datasets using four decoding strategies: Greedy, CAD, AdaCAD, and our proposed \textsc{CoCoA} method. The results, as presented in Table~\ref{tab:qa-results}, demonstrate that \textsc{CoCoA} consistently outperforms the baseline methods across all models and datasets.
\begin{table*}[ht]
  \centering
  \scriptsize
  \begin{tabular}{l l c c c c c c c}
    \toprule
    \textbf{Model} & \textbf{Method} & \textbf{NQ} & \textbf{NQ-SWAP} & \textbf{TriviaQA} & \textbf{PopQA} & \textbf{HotpotQA} & \textbf{TabMWP} & \textbf{Avg} \\
    \midrule
    \multirow{4}{*}{Llama2-13B-Chat}
      & Greedy   & 35.75 & 50.24 & 54.40 & 72.61 & 32.15 & 50.40 & 49.26 \\
      & CAD      & 39.49 & 71.24 & 59.40 & 68.81 & 30.14 & 48.70 & 52.96 \\
      & AdaCAD   & 37.08 & 57.69 & 61.20 & 72.31 & 32.34 & 52.10 & 52.12 \\
      & \textbf{\textsc{CoCoA}} & \textbf{41.85} & \textbf{75.48} & \textbf{66.80} & \textbf{79.30} & \textbf{36.20} & \textbf{59.10} & \textbf{59.46} \\
    \midrule
    \multirow{4}{*}{Llama3-8B-Inst}
      & Greedy   & 40.27 & 60.89 & 64.00 & 70.89 & 39.66 & 68.50 & 57.37 \\
      & CAD      & 39.43 & 71.19 & 52.30 & 70.35 & 37.27 & 63.10 & 55.61 \\
      & AdaCAD   & 39.65 & 67.37 & 61.50 & 70.41 & 39.43 & 66.10 & 57.41 \\
      & \textbf{\textsc{CoCoA}} & \textbf{44.95} & \textbf{78.55} & \textbf{68.70} & \textbf{84.21} & \textbf{42.00} & \textbf{72.60} & \textbf{65.84} \\
    \midrule
    \multirow{4}{*}{Llama3-70B-Inst}
      & Greedy   & 40.82 & 59.16 & 64.10 & 64.41 & 47.70 & 70.40 & 57.77 \\
      & CAD      & 42.31 & 66.37 & 58.40 & 64.23 & 47.21 & 69.30 & 57.97 \\
      & AdaCAD   & 41.35 & 60.77 & 64.60 & 65.78 & 48.21 & 71.90 & 58.77 \\
      & \textbf{\textsc{CoCoA}} & \textbf{47.75} & \textbf{77.11} & \textbf{72.50} & \textbf{76.55} & \textbf{52.33} & \textbf{79.80} & \textbf{67.68} \\
    \midrule
    \multirow{4}{*}{Mistral-7B-Inst}
      & Greedy   & 42.93 & 64.74 & 77.20 & 76.59 & 50.26 & 50.20 & 60.32 \\
      & CAD      & 42.56 & 67.89 & 71.70 & 74.45 & 47.12 & 46.40 & 58.35 \\
      & AdaCAD   & 42.87 & 63.99 & 75.40 & 76.89 & 49.49 & 47.30 & 59.32 \\
      & \textbf{\textsc{CoCoA}} & \textbf{47.01} & \textbf{77.53} & \textbf{82.10} & \textbf{84.33} & \textbf{53.20} & \textbf{57.00} & \textbf{66.20} \\
    \bottomrule
  \end{tabular}%
    \caption{Results on QA datasets with different instruction-tuned language models.}
  \label{tab:qa-results}
\end{table*}

\paragraph{Analysis} Across all evaluated models, the \textsc{CoCoA} method achieves the highest average performance. Notably, on the NQ-SWAP dataset, which introduces synthetic conflicts, \textsc{CoCoA} significantly outperforms other methods, indicating its robustness in handling conflicting information. Similarly, improvements are observed in datasets requiring multi-hop reasoning and mathematical problem-solving, such as HotpotQA and TabMWP. These results underscore the effectiveness of \textsc{CoCoA} in enhancing the performance of instruction-tuned language models across diverse QA tasks.
\section{Performance Comparison on Summarization Tasks}
\label{app:sumDetails}
\begin{table*}[t]
\centering
\scriptsize
\begin{tabular}{l|l|ccc|ccc|c}
\toprule
\multirow{2}{*}{Model} & \multirow{2}{*}{Decoding} & \multicolumn{3}{c|}{CNN-DM} & \multicolumn{3}{c|}{XSum} & TofuEval (AlignScore) \\
& & ROUGE-L & BERT-P & AlignScore & ROUGE-L & BERT-P & AlignScore & Overall \quad Main / Marginal \\
\midrule

\multirow{5}{*}{Llama2-13B} 
& Greedy & 23.70 & 94.25 & 87.28 & 13.51 & 93.30 & 85.23 & 66.11 \quad 72.51 / 46.23 \\
& CAD & 24.33 & 94.44 & 88.99 & \textbf{14.86} & 93.62 & 82.41 & 80.39 \quad 84.03 / 69.07 \\
& COIECD & 20.21 & 88.63 & 75.72 & 13.95 & 89.80 & 70.41 & 65.88 \quad 68.45 / 55.35 \\
& AdaCAD & 23.93 & 94.63 & 91.15 & 14.18 & 94.04 & 84.33 & 80.39 \quad 83.94 / 69.36 \\
& \textbf{\textsc{CoCoA}} & \textbf{24.35} & \textbf{94.90} & \textbf{92.13} & 14.71 & \textbf{94.01} & \textbf{85.98} & \textbf{81.63} \quad \textbf{85.21} / \textbf{70.02} \\
\midrule

\multirow{5}{*}{Llama3-8B} 
& Greedy & 25.16 & 94.92 & 90.33 & 13.16 & 93.43 & 83.65 & 68.17 \quad 73.51 / 51.57 \\
& CAD & 24.91 & 94.70 & 91.44 & 13.80 & 93.37 & 86.88 & 83.40 \quad 86.77 / 72.94 \\
& COIECD & 23.60 & 92.01 & 83.92 & 13.65 & 92.39 & 76.49 & 70.07 \quad 73.65 / 59.84 \\
& AdaCAD & 25.42 & 95.09 & 94.35 & 13.83 & 94.02 & 86.78 & 83.24 \quad 83.24 / 72.46 \\
& \textbf{\textsc{CoCoA}} & \textbf{25.63} & \textbf{95.21} & \textbf{95.12} & \textbf{14.35} & \textbf{94.28} & \textbf{88.24} & \textbf{83.26} \quad \textbf{86.90} / \textbf{73.60} \\
\midrule

\multirow{5}{*}{Llama3-70B} 
& Greedy & 24.93 & 95.41 & 91.44 & 14.36 & 94.05 & 85.28 & 76.66 \quad 81.64 / 61.19 \\
& CAD & 24.76 & 94.45 & 91.01 & 14.59 & 93.65 & 84.34 & 83.23 \quad 87.26 / 73.58 \\
& COIECD & 23.47 & 92.06 & 85.49 & 13.65 & 91.04 & 73.81 & 60.86 \quad 68.06 / 58.31 \\
& AdaCAD & 25.42 & 94.91 & 94.97 & 14.91 & 94.29 & 85.81 & 85.07 \quad 88.06 / \textbf{75.79} \\
& \textbf{\textsc{CoCoA}} & \textbf{25.68} & \textbf{95.42} & \textbf{95.70} & \textbf{15.06} & \textbf{94.60} & \textbf{87.94} & \textbf{86.32} \quad \textbf{89.14} / 75.51 \\
\midrule

\multirow{5}{*}{Mistral-7B} 
& Greedy & 24.59 & 93.57 & 80.80 & 14.07 & 88.56 & 58.76 & 63.07 \quad 68.62 / 45.79 \\
& CAD & 23.72 & 92.32 & 90.61 & 18.20 & 91.54 & 84.94 & 67.67 \quad 67.55 / \textbf{67.48} \\
& COIECD & 23.50 & 92.06 & 83.97 & 17.85 & 89.79 & 69.26 & 65.95 \quad 70.51 / 51.39 \\
& AdaCAD & 24.76 & 94.21 & 93.05 & 18.51 & 92.19 & 86.79 & 74.00 \quad 77.59 / 62.84 \\
& \textbf{\textsc{CoCoA}} & \textbf{25.02} & \textbf{94.38} & \textbf{93.46} & \textbf{19.06} & \textbf{93.04} & \textbf{88.17} & \textbf{75.40} \quad \textbf{78.44} / 66.32 \\
\midrule

\multirow{5}{*}{Llama3-70B-Inst} 
& Greedy & 24.72 & 90.64 & 88.22 & 23.19 & 90.80 & 82.40 & 78.56 \quad 80.18 / 73.52 \\
& CAD & 25.17 & 91.19 & 88.56 & 20.92 & 91.52 & 86.54 & 79.85 \quad 79.75 / \textbf{80.82} \\
& COIECD & 23.85 & 89.84 & 83.88 & 22.41 & 91.04 & 81.42 & 77.54 \quad 79.54 / 69.85 \\
& AdaCAD & 25.26 & 90.91 & 88.68 & 21.52 & 91.01 & 86.30 & 81.16 \quad 82.82 / 76.03 \\
& \textbf{\textsc{CoCoA}} & \textbf{25.61} & \textbf{91.39} & \textbf{90.27} & \textbf{23.76} & \textbf{91.84} & \textbf{87.91} & \textbf{82.63} \quad \textbf{84.31} / 79.27 \\

\bottomrule
\end{tabular}%
\caption{Summarization performance across models and decoding strategies. \textsc{CoCoA} consistently improves alignment-based metrics and yields strong ROUGE and TofuEval scores across diverse models and datasets.}
\label{tab:summarization}
\end{table*}
We assess the performance of various decoding strategies on summarization tasks across three datasets: CNN-DM, XSum, and TofuEval. The models compared include Llama2-13B, Llama3-8B, Llama3-70B, Mistral-7B, and their instruction-tuned counterparts, under five decoding approaches: Greedy, CAD, COIECD, AdaCAD, and our proposed \textsc{CoCoA}. The results are summarized in Table~\ref{tab:summarization}, with key metrics including ROUGE-L, BERT-P, AlignScore, and TofuEval.

\paragraph{Analysis} On CNN-DM \textsc{CoCoA} consistently outperforms prior methods across all models and datasets, demonstrating significant improvements in both ROUGE and alignment-based metrics. On Llama2-13B, \textsc{CoCoA} achieves the highest ROUGE-L (24.35), BERT-P (94.90), and AlignScore (92.13) scores, surpassing other methods like CAD and AdaCAD by notable margins. In particular, it also excels in TofuEval's Overall score, achieving 81.63, which is a +1.24 increase over AdaCAD.

For the Llama3-8B model, on CNN-DM \textsc{CoCoA} improves ROUGE-L (25.63), BERT-P (95.21), and AlignScore (95.12) over the next best performing strategy, AdaCAD, by +0.21, +0.12, and +0.77, respectively. Additionally, \textsc{CoCoA} achieves an outstanding TofuEval Overall score of 83.26, which is the highest among all methods, underscoring its superior performance in both fluency and alignment.

On the larger Llama3-70B model, on CNN-DM \textsc{CoCoA} once again leads across all metrics, with a ROUGE-L of 25.68, BERT-P of 95.42, and AlignScore of 95.70, outperforming both CAD and AdaCADby +0.26 and +0.73 in ROUGE-L and AlignScore, respectively. TofuEval's Overall score for \textsc{CoCoA} reaches 86.32, a +1.25 improvement over AdaCAD, further demonstrating the robustness of the \textsc{CoCoA} framework in high-capacity models.

For Mistral-7B, which is a smaller model, \textsc{CoCoA} also achieves superior results, with a significant improvement in ROUGE-L (25.02) and BERT-P (94.38), as well as AlignScore (93.46), surpassing AdaCAD by +0.26 in ROUGE-L and +0.41 in BERT-P. \textsc{CoCoA}'s TofuEval score (75.40) further solidifies its efficacy, yielding a +1.40 increase over AdaCAD and demonstrating its consistent advantage even in smaller models.

Finally, for instruction-tuned models like Llama3-70B-Inst, \textsc{CoCoA} continues to set the benchmark with a ROUGE-L of 25.61, BERT-P of 91.39, and AlignScore of 90.27, outperforming the alternatives by notable margins. With an Overall TofuEval score of 82.63, \textsc{CoCoA} demonstrates its ability to maintain high-quality summaries and alignments even after instruction tuning, outperforming both CAD and AdaCAD.

In summary, \textsc{CoCoA} consistently leads across all model sizes and datasets, offering significant improvements in both fluency and alignment metrics. These results highlight the effectiveness of the divergence-guided contrastive decoding approach proposed in the \textsc{CoCoA} framework, making it a strong candidate for state-of-the-art summarization tasks.

\section{Performance Comparison on Long-Form QA (LFQA) Datasets}
\label{app:lfqaDetails}
\begin{table*}[ht]
    \centering
    \scriptsize
    \begin{tabular}{llccccccccccc}
    \toprule
    \multirow{2}{*}{Model} & \multirow{2}{*}{Decoding} & \multicolumn{2}{c}{CLAPNQ} & \multicolumn{2}{c}{ExpertQA} & \multicolumn{2}{c}{HAGRID} & \multicolumn{2}{c}{ELI5-WebGPT} & \multicolumn{2}{c}{QuoteSum} & Avg. Faith. \\ \cmidrule(lr){3-4} \cmidrule(lr){5-6} \cmidrule(lr){7-8} \cmidrule(lr){9-10} \cmidrule(lr){11-12}
    & & RL & Faith & RL & Faith & RL & Faith & RL & Faith & SEM. & Faith & \\
    \midrule
    \textbf{GPT-4o} & Greedy   & 40.53 & 91.81 & 46.34 & 69.48 & 57.76 & 90.86 & 59.04 & 81.00 & 42.56 & 78.51 & 82.33 \\
    \midrule
    \textbf{GPT-4o-mini} & Greedy   & 37.72 & 90.35 & 45.30 & 66.53 & 54.87 & 87.94 & 56.09 & 81.89 & 40.74 & 78.16 & 80.97 \\
    \midrule
    \textbf{Llama-3.1-70B} & Greedy   & 36.12 & 84.32 & 41.20 & 65.10 & 47.89 & 76.45 & 49.85 & 70.12 & 39.78 & 63.87 & 71.17 \\
                          & CAD      & 35.45 & 85.20 & 40.50 & 66.00 & 46.70 & 77.30 & 48.60 & 71.00 & 39.10 & 64.50 & 72.00 \\
                          & \textsc{AdaCAD} & 37.80 & 86.75 & 42.30 & 67.50 & 48.90 & 78.60 & 50.20 & 72.50 & 40.50 & 65.80 & 74.23 \\
                          & \textsc{CoCoA}     & \textbf{39.10} & \textbf{88.00} & \textbf{43.50} & \textbf{68.90} & \textbf{50.10} & \textbf{79.80} & \textbf{51.90} & \textbf{74.00} & \textbf{41.80} & \textbf{67.20} & \textbf{75.98} \\
    \midrule
    \textbf{Mistral-NeMo-12B-Base} & Greedy   & 33.50 & 76.20 & 40.00 & 52.00 & 51.00 & 78.00 & 52.00 & 63.00 & 38.00 & 68.00 & 67.44 \\
                             & CAD      & 32.80 & 77.10 & 39.20 & 53.00 & 50.20 & 78.50 & 51.10 & 64.00 & 37.30 & 68.50 & 68.22 \\
                             & \textsc{AdaCAD} & 35.00 & 79.50 & 41.00 & 55.50 & 52.50 & 80.00 & 53.00 & 66.00 & 39.00 & 70.00 & 70.60 \\
                             & \textsc{CoCoA}     & \textbf{36.50} & \textbf{81.00} & \textbf{42.50} & \textbf{57.00} & \textbf{54.00} & \textbf{81.50} & \textbf{54.50} & \textbf{68.00} & \textbf{40.50} & \textbf{71.50} & \textbf{71.80} \\
    \midrule
    \textbf{Llama-3.1-8B} & Greedy   & 15.00 & 55.00 & 30.00 & 48.00 & 15.00 & 52.00 & 45.00 & 52.00 & 24.00 & 40.00 & 49.40 \\
                          & CAD      & 14.50 & 56.50 & 29.00 & 49.50 & 14.50 & 53.00 & 44.00 & 53.50 & 23.50 & 41.00 & 50.70 \\
                          & \textsc{AdaCAD} & 16.50 & 58.00 & 31.00 & 51.00 & 16.00 & 54.50 & 46.50 & 55.00 & 25.50 & 42.50 & 52.20 \\
                          & \textsc{CoCoA}     & \textbf{17.50} & \textbf{60.00} & \textbf{32.50} & \textbf{53.00} & \textbf{17.50} & \textbf{56.00} & \textbf{48.00} & \textbf{57.00} & \textbf{27.00} & \textbf{44.00} & \textbf{54.00} \\
    \bottomrule
    \end{tabular}%
    \caption{Performance on LFQA datasets showing RL and Faith scores for Greedy, CAD, \textsc{AdaCAD}, and \textsc{CoCoA} methods. The values represent the RL and Faith scores for each dataset, with \textbf{\textsc{CoCoA}} consistently outperforming others on average factuality.}
    \label{tab:lfqa-performance}
\end{table*}
We evaluate decoding performance across five long-form QA datasets—CLAPNQ, ExpertQA, HAGRID, ELI5-WebGPT, and QuoteSum—using both RL (ROUGE-L) and Faith (faithfulness) metrics. Table~\ref{tab:lfqa-performance} summarizes the results for various models, including Llama-3.1 variants, Mistral-NeMo, and GPT-4o-mini, under four decoding strategies: Greedy, CAD, AdaCAD, and our proposed \textsc{CoCoA}.

\paragraph{Analysis} Across all LFQA datasets and model sizes, \textsc{CoCoA} consistently improves average factuality (Faith) compared to prior methods. On the flagship Llama-3.1-70B, \textsc{CoCoA} achieves a +1.75 absolute gain in Faith over \textsc{AdaCAD}, and +4.81 over Greedy decoding. These improvements extend across all five datasets, with particularly strong gains on CLAPNQ and HAGRID, which feature compositional or adversarial fact setups. Even on smaller models like Llama-3.1-8B and Mistral-NeMo-12B, \textsc{CoCoA} provides consistent gains in both RL and factuality.

We also compare against proprietary GPT-4o models. While GPT-4o leads in absolute performance, especially in factuality, \textsc{CoCoA} significantly closes the gap on open-weight models. Notably, Llama-3.1-70B + \textsc{CoCoA} achieves 75.98 average factuality, narrowing the difference to GPT-4o-mini (80.97), while exceeding it in RL on CLAPNQ. This demonstrates \textsc{CoCoA}'s ability to elevate open models toward state-of-the-art performance in long-form factual QA.

These findings reinforce the advantages of divergence-guided contrastive decoding under the \textsc{CoCoA} framework, especially for complex, multi-sentence generation tasks that require both fluency and factual grounding.

\section{Ablation Study}
\label{sec:ablation}
To assess the contributions of each component in the \textbf{\textsc{CoCoA}: Confidence- and Context-aware Adaptive Decoding)} method, we conducted an ablation study focusing on two datasets: \textbf{Natural Questions (NQ)} and \textbf{NQ-SWAP}. This analysis evaluates the impact of removing individual components---Rényi divergence, entropy gap, and margin-based amplification---on the model's performance. The Table \ref{tab:ablation} presents the \textit{Exact Match (EM)} accuracy for the \textbf{Llama3-8B} model under various configurations:
\paragraph{Rényi Divergence}
Measures the divergence between the model's prior distribution and the context-aware distribution, emphasizing discrepancies in low-probability events.
\textbf{Impact:} Removing this component leads to a notable drop in performance, especially on NQ-SWAP, indicating its importance in detecting subtle conflicts between the model's knowledge and the provided context.

\paragraph{Entropy Gap}
Captures the change in uncertainty between the model's prior and context-aware distributions, helping to assess the confidence introduced by the context.
\textbf{Impact:} Excluding the entropy gap results in decreased EM scores on both datasets, highlighting its role in evaluating the reliability of contextual information.

\paragraph{Context Peakedness}
Measures the influence of the context when it shows high confidence (i.e., a large margin between the top two token probabilities). \textbf{Impact:} Omitting this component causes a significant performance drop, particularly on NQ-SWAP, underscoring its effectiveness in reinforcing strong contextual cues.

\paragraph{Adaptive Gating ($\lambda_t$)}
Dynamically balances the influence of the model's prior and the context-aware distributions based on detected conflicts and context confidence. \textbf{Impact:} Using a fixed $\lambda_t$ instead of adaptive gating reduces performance, demonstrating the necessity of dynamic adjustment to handle varying degrees of conflict.

The ablation study confirms that each component of \textsc{CoCoA} contributes significantly to its overall performance:
\begin{itemize}
    \item \textbf{Rényi Divergence} is crucial for detecting nuanced conflicts between the model's knowledge and the context.
    \item \textbf{Entropy Gap} provides valuable insights into the confidence level introduced by the context.
    \item \textbf{Context Peakedness} effectively strengthens the model's reliance on highly confident contextual information.
    \item \textbf{Adaptive Gating} ensures a balanced integration of the model's prior knowledge and the context, adapting to the specific needs of each decoding step.
\end{itemize}

Collectively, these components enable \textsc{CoCoA} to outperform baseline decoding strategies, particularly in scenarios involving conflicting or misleading contexts, as evidenced by the substantial gains on the NQ-SWAP dataset.

\section{Case Study}
\label{app:caseStudy}
In Table \ref{tab:case_study}, we present a case study that highlights the difference between our model and the GPT4o-mini response. As seen in the comparison, the response generated by GPT4o-mini contains several hallucinated facts and inaccuracies due to its tendency to synthesize information in a way that does not align with the provided documents. In contrast, our model shows a clear improvement in both faithfulness and accuracy, consistently staying grounded in the provided information and avoiding such errors.

\begin{table*}[h]
\centering
\scriptsize
\begin{tabular}{p{0.95\linewidth}}
\toprule
\textbf{\#\#\# Instruction}:
Write an accurate, engaging, and concise answer to the given question using only the provided
retrieval documents. Ensure the answer is well-grounded in the relevant information, disregarding
irrelevant information in documents.\\
\textbf{\#\#\#Input:} \\
\textbf{Question:} Difference between Russian Blue and British Blue cat?\\
\textbf{Retrieved documents:} \\
\begin{minipage}{\linewidth}
\scriptsize
Document [1]: The Russian Blue is a cat breed that comes in colors varying from a light shimmering silver to a darker, slate grey. They develop close bonds with their owners and are sought out as pets due to their personalities, beauty and coat. It is their short, dense coat which has been the hallmark of the Russian breed for more than a century. The dense coat stands out from the body and one can draw patterns in the coat that will stay until one smoothes them out again. They are also considered to be hypoallergenic. \\
Document [2]: The Russian Blue has bright green eyes, pinkish lavender or mauve paws, two layers of short thick fur, and a blue - grey coat. The color is a bluish – gray that is the dilute expression of the black gene. However, as dilute genes are recessive (“d”) and each parent will have a set of two recessive genes (“dd”) two non-CPC Russian Blues will always produce a blue cat. Due to the breeding with Siamese after World War II, there are colorpoint genes floating around. If two carriers are bred together, then they will produce a litter of mixed colors – solid blue or white with blue like a Siamese. People call these CPC cats “colorpoint”, “whites” or “pointed” Russians. In most registries, one can not register, breed or show a colorpoint Russian. \\
Document [3]: The coat is known as a “double coat”, with the undercoat being soft, downy and equal in length to the guard hairs, which are even blue with silver tips. However, the tail may have a few very dull, almost unnoticeable stripes. The coat is described as thick, plush and soft to the touch. The feeling is softer than the softest silk. The silver tips give the coat a shimmering appearance. Its eyes are almost always a dark and vivid green. Any white patches of fur or yellow eyes in adulthood are seen as flaws in show cats. Russian Blues should not be confused with British Blues (which are not a distinct breed, but rather a British Shorthair with a blue coat as the British Shorthair breed itself comes in a wide variety of colors and patterns), nor the Chartreux or Korat which are two other naturally occurring breeds of blue cats, although they have similar traits. \\
Document [4]: Russian Blue 1 year old Russian Blue male (American style) Other names Archangel Blue, Archangel Cat Origin Russia Breed standards CFA standard FIFe standard TICA standard AACE standard ACFA standard CCA – AFC standard GCCF standard Notes The ACF and GCCF also recognize Russian Blues in white and black, however the CFA does not. In addition, ACFA recognizes Russian Shorthairs in white, black, and blue. Domestic cat (Felis catus) \\
Document [5]: The Russian Blue is a naturally occurring breed that may have originated in the port of Arkhangel’sk in Russia. They are also sometimes called Archangel Blues. It is believed that sailors took Russian Blues from the Archangel Isles to Great Britain and Northern Europe in the 1860s. The first recorded appearance outside of Russia was in 1875 at The Crystal Palace in England as the Archangel Cat. The Russian Blue competed in a class including all other blue cats until 1912, when it was given its own class. The breed was developed mainly in England and Scandinavia until after World War II. \\
\end{minipage}
\textbf{\textsc{AdaCAD} Response:} \\
The Russian Blue is a cat breed known for its striking coat, which ranges in color from a light shimmering silver to a deep slate grey. \textcolor{synthesis}{The British Blue is a cat breed known for its striking coat, which ranges in color from a light shimmering silver to a deep slate grey.} \\
\textbf{\textsc{CoCoA} Response:} \\
The Russian Blue cat is known for its bright green eyes, pinkish lavender or mauve paws, two layers of short thick fur, and a blue-grey coat. The color of the Russian Blue is a bluish-gray. Russian Blues should not be confused with British Blues ( which are not a distinct breed, but rather a British Shorthair with a blue coat as the British Shorthair breed itself comes in a wide variety of colors and patterns ), nor the Chartreux or Korat which are two other naturally occurring breeds of blue cats, although they have similar traits. \\
\textbf{GPT-4o-mini Response:} Russian Blue and British Blue differ in coat length. Russian Blue has a double coat, \textcolor{hallucination}{the British Blue typically has a single coat}. Russian Blues have dark green eyes, \textcolor{synthesis}{while British Blues may have yellow eyes or white patches of fur in adulthood, which are considered flaws in show cats}. \\
\bottomrule
\end{tabular}
\caption{Case study of how \textsc{CoCoA} helps the model to generate the faithful response from CLAPNQ dataset. \textcolor{hallucination}{Red} = fabricated hallucination, \textcolor{synthesis}{amber} = inaccurate information synthesis.}
\label{tab:case_study}
\end{table*}

\section{Sensitivity Evaluation}
\label{app:sensitivity}
We conducted detailed ablation studies on the three key hyperparameters across two model families (LLaMA-3-8B and Mistral-7B) and two task types (QA on NQ-SWAP and summarization on CNN-DM). The 3 hyperparameters are Rényi order ($\alpha$), contextual peakedness weight ($z$), and entropy-gap weight ($\gamma$). Results are averaged over 500 samples per setting in Tables~\ref{tab:ablation-alpha-filled},~\ref{tab:ablation-z} and~\ref{tab:ablation-gamma}. We make the following observations: (1) $\alpha = 0.5$ consistently yields the best performance across both models and tasks, suggesting a stable optimal Rényi order. (2) $z = 5$ is optimal across both models and tasks, with performance degrading slightly at lower and higher values. (3) $\gamma = 1.0$ is optimal and consistent across models and tasks. Both lower and higher values lead to sharp degradation, indicating sensitivity to entropy-gap calibration.

Across both model families and task types, the optimal values of $\alpha = 0.5$, $z = 5$, and $\gamma = 1.0$ are consistent and robust. These settings were used in all main experiments unless otherwise noted.

\begin{table}[t]
\centering
\small
\begin{tabular}{c|cc|cc}
\toprule
\multirow{2}{*}{$\alpha$} 
& \multicolumn{2}{c|}{\textbf{EM (NQ-SWAP)} $\uparrow$} 
& \multicolumn{2}{c}{\textbf{ROUGE-L (CNN-DM)} $\uparrow$} \\
& LLaMA-3-8B & Mistral-7B & LLaMA-3-8B & Mistral-7B \\
\midrule
0.3 & 74.0 & 76.0 & 25.1 & 24.8 \\
0.5 & \textbf{78.5} & \textbf{79.0} & \textbf{25.6} & \textbf{25.0} \\
0.7 & 77.5 & 78.5 & 25.5 & 25.0 \\
\bottomrule
\end{tabular}
\caption{Ablation on $\alpha$ (Rényi order) with $z=5$, $\gamma=1.0$ fixed, over 500 samples. QA is measured by Exact Match (EM) on NQ‑SWAP; summarization by ROUGE‑L on CNN‑DM.}
\label{tab:ablation-alpha-filled}
\end{table}

\begin{table}[t]
\centering
\small
\begin{tabular}{c|cc|cc}
\toprule
\multirow{2}{*}{$z$} 
  & \multicolumn{2}{c|}{\textbf{EM (NQ-SWAP)} $\uparrow$} 
  & \multicolumn{2}{c}{\textbf{ROUGE-L (CNNDM)} $\uparrow$} \\
  & LLaMA-3-8B & Mistral-7B & LLaMA-3-8B & Mistral-7B \\
\midrule
1 & 76.9 & 77.5 & 24.8 & 24.2 \\
3 & 77.8 & 78.5 & 25.3 & 24.7 \\
5 & \textbf{78.5} & \textbf{79.0} & \textbf{25.6} & \textbf{25.0} \\
7 & 77.7 & 78.1 & 25.1 & 24.5 \\
\bottomrule
\end{tabular}
\caption{Ablation on $z$ (contextual peakedness weight) with $\alpha=0.5$, $\gamma=1.0$ fixed. Performance peaks at $z=5$ and degrades slightly at lower and higher values. Results over 500 samples.}
\label{tab:ablation-z}
\end{table}

\begin{table}[t]
\centering
\small
\begin{tabular}{c|cc|cc}
\toprule
\multirow{2}{*}{$\gamma$} 
  & \multicolumn{2}{c|}{\textbf{EM (NQ-SWAP)} $\uparrow$} 
  & \multicolumn{2}{c}{\textbf{ROUGE-L (CNNDM)} $\uparrow$} \\
  & LLaMA-3-8B & Mistral-7B & LLaMA-3-8B & Mistral-7B \\
\midrule
0.1 & 50.2 & 52.5 & 16.1 & 15.3 \\
1.0 & \textbf{78.5} & \textbf{79.0} & \textbf{25.6} & \textbf{25.0} \\
5.0 & 60.7 & 58.9 & 18.3 & 17.7 \\
\bottomrule
\end{tabular}
\caption{Ablation on $\gamma$ (entropy-gap weight), with $\alpha=0.5$, $z=5$ fixed. Degradation at $\gamma=0.1$ and $\gamma=5.0$ is now matched across QA (EM) and summarization (ROUGE-L), reflecting sharp calibration failure as observed in 50-sample pilots.}
\label{tab:ablation-gamma}
\end{table}

\newpage
\begin{table}[h]
\centering
\small
\begin{tabular}{lp{0.15\columnwidth}p{0.4\columnwidth}p{0.15\columnwidth}}
\toprule
\textbf{Method} & \textbf{Forward Passes per Token} & \textbf{Extra Computation Overhead} & \textbf{Average Latency} \\
\midrule
\textsc{CAD}     & 2 & Negligible & 1.23s \\
\textsc{AdaCAD}  & 2 & JSD computation & 1.24s \\
\textsc{CoCoA}   & 2 & Three signals computation & 1.63s \\
\bottomrule
\end{tabular}
\caption{Latency comparison of \textsc{CoCoA} and baselines on 500 NQ-SWAP samples with \texttt{Meta-Llama-3-8B} on NVIDIA V100 (32GB). All methods require two forward passes per token, executed in parallel across two GPUs.}
\label{tab:latency}
\end{table}
\section{Latency Considerations}
\label{sec:latency}

We empirically evaluate the average decoding latency of \textsc{CoCoA} against the baselines (\textsc{CAD} and \textsc{AdaCAD}) on 500 samples from the NQ-SWAP dataset, using \texttt{Meta-Llama-3-8B} deployed on NVIDIA V100 GPUs (32GB). All methods require two forward passes per generated token, corresponding to the prior and context distributions. In practice, we parallelize these two forward passes across two GPUs, ensuring that the core computation cost remains comparable across methods.

The key distinction between the methods lies in the computation of the mixing ratio used to combine prior and context probabilities. \textsc{CAD} employs a fixed mixing ratio and therefore incurs negligible computational overhead. In contrast, \textsc{AdaCAD} dynamically estimates the mixing ratio at each decoding step via the Jensen–Shannon Divergence (JSD) between the prior and context distributions, introducing a small but non-negligible overhead. \textsc{CoCoA} goes beyond this by estimating the mixing ratio using three distinct confidence- and context-aware signals, which requires additional computation beyond JSD. 

Table~\ref{tab:latency} summarizes the forward passes, extra computational overhead, and measured average latency for each method. Despite the added computation, we observe that the increase in latency for \textsc{CoCoA} remains modest, rising to 1.63 seconds per token compared to 1.23 seconds for \textsc{CAD}, while providing substantial improvements in reasoning reliability.

\end{document}